\documentclass[10pt,twocolumn,letterpaper]{article}

\usepackage{wacv}              % To produce the CAMERA-READY version
\definecolor{wacvblue}{rgb}{0.21,0.49,0.74}
\usepackage[pagebackref,breaklinks,colorlinks,allcolors=wacvblue]{hyperref}

\usepackage{amsmath} % assumes amsmath package installed
\usepackage{amssymb}  % assumes amsmath package installed

\usepackage{graphicx}
\usepackage{algorithm}
\usepackage{algorithmic}
\usepackage{multirow}
\usepackage{adjustbox}
\usepackage{booktabs} 
 \usepackage{hyperref}
\usepackage{makecell} % in your preamble
\usepackage{float}
\usepackage{amsmath}
\usepackage{subcaption}

\def\wacvPaperID{1486} % *** Enter the WACV Paper ID here
\def\confName{WACV}
\def\confYear{2027}

\title{3D-BBDA: Backpropagation-Free Black-Box Domain Adaptation for Robust 3D Point Cloud Recognition}

\author{Shivangi Rai\textsuperscript{*}, Kunal Jangid\textsuperscript{*}, Vaibhav Kumar\textsuperscript{\dag}, Vinod K Kurmi\\
Department of Data Science and Engineering, \\ Indian Institute of Science Education and Research Bhopal, India\\
{\tt\small \{shivangir23, kunal24, vaibhav, vinodkk\}@iiserb.ac.in}\\
}

\begin{document}
\maketitle
\begingroup
\renewcommand{\thefootnote}{}
\footnotetext{%
\textsuperscript{*}Equal contribution.
\hspace{1.5em}
\textsuperscript{\dag}Corresponding author.%
}
\endgroup
\begin{abstract}
Robust 3D point-cloud recognition is critical for robotics, mapping, and urban sensing, yet point-cloud classifiers often suffer performance drops under domain shifts caused by sensor changes, environmental variation, and acquisition artifacts. Many existing adaptation methods are difficult to deploy in practice because they require access to source data or the internal parameters of the source model, which may be restricted by privacy, licensing, or security constraints. We present \emph{3D-BBDA}, a \emph{backpropagation-free black-box domain adaptation} framework that adapts a source-trained predictor to unlabeled target data using only black-box outputs. Our method performs pseudo-label guided \emph{prototype-based alignment} and employs a \emph{weighted prediction} module to fuse source logits with alignment logits, reducing sensitivity to noisy pseudo-labels. To stabilize prototype memory when pseudo-label coverage is incomplete, we construct a \emph{Gaussian-perturbed prototype prior} from unlabeled target predictions and use it as a fallback for missing or unreliable class prototypes during alignment.
Experiments on four corrupted 3D benchmarks demonstrate consistent improvements over strong baselines. On ModelNet-40C, 3D-BBDA achieves mean errors of 32.08\% (PointNet) and 29.62\% (GDANet), outperforming Source-only (35.88\% / 34.03\%) and BFTT3D (33.78\% / 32.36\%). 3D-BBDA requires only a single forward pass with no gradient updates making it suitable for real-time, privacy-preserving deployment in transferable 3D recognition pipelines. Project page is available at \url{https://visdomlab.github.io/3D_BBDA}
\end{abstract}
    
\section{Introduction}
\label{sec:intro}

Robust 3D perception is fundamental to autonomous systems, including mobile robots, autonomous vehicles, unmanned aerial platforms, and urban mapping solutions. These systems rely heavily on point cloud data acquired from LiDAR, depth cameras, and other 3D sensors for object recognition, navigation, and interaction in unstructured environments. In real deployments, perception models often operate under changing environmental conditions, sensor variations, and acquisition artifacts that induce domain shifts, leading to significant performance degradation \cite{qin2019pointdan,ren2022uni3da}. Ensuring robustness under such distributional shifts is critical for the safe and reliable deployment of AI-driven robotic systems.

While unsupervised domain adaptation (UDA) attempts to transfer knowledge from labeled source domains to unlabeled target domains \cite{ganin2016domain,tzeng2017adversarial}, most UDA methods require access to source data or source model parameters during adaptation \cite{liang2022dine}. In practical robotic deployments, this assumption is often unrealistic. Perception models may be proprietary, certified, or frozen after training, and source datasets may be inaccessible due to privacy, storage, or licensing constraints \cite{rai2025label}. Moreover, online gradient-based fine-tuning is frequently infeasible on edge robotic platforms due to computational and safety considerations. These limitations motivate \emph{black-box} and \emph{source-free} adaptation strategies that rely on inference-time observations.

Black-Box Domain Adaptation (BBDA) addresses this setting by treating the source model as an immutable predictor and adapting only through its outputs \cite{liang2020we,liang2022dine,li2024comprehensive}. However, existing BBDA research has primarily focused on 2D vision tasks. Extending BBDA to 3D point clouds is non-trivial due to unordered structures, irregular spatial distributions, and varying density patterns. Furthermore, existing approaches often depend on backpropagation or source statistics for alignment, limiting their applicability in strictly constrained deployment scenarios.
These highlight a critical research gap, necessitating novel approaches tailored to the unique characteristics of 3D point clouds under black-box constraints.
Additionally, these methods typically enforce the use of the same network architecture as the source model, which restricts the flexibility to adopt different architectures for the target model during adaptation \cite{liang2022dine, zhang2023black}. This limitation hinders the ability to tailor target networks to specific tasks or computational constraints. 
\begin{figure}[t]
    \centering
    \includegraphics[width=0.75\columnwidth]{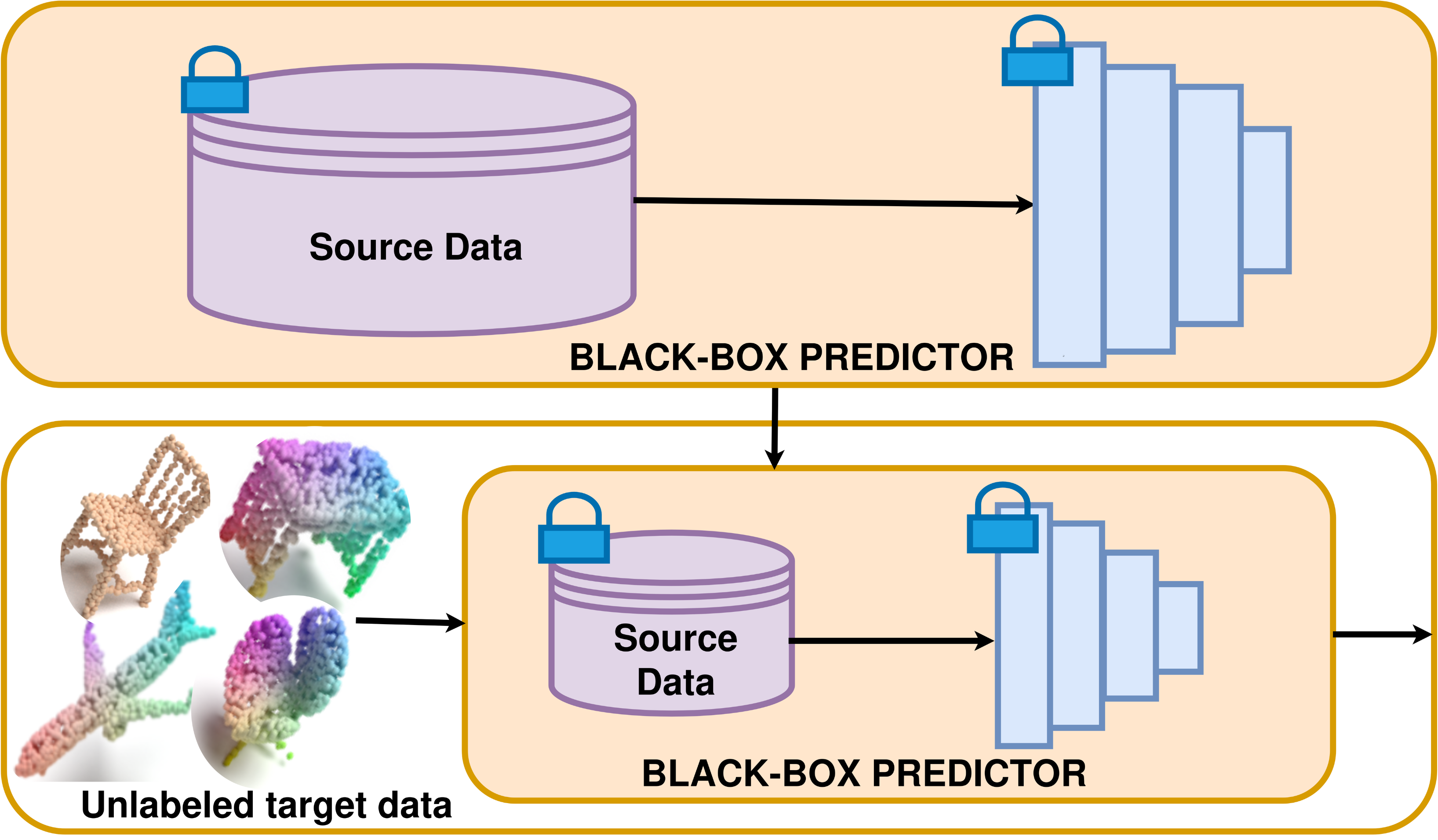}
    \caption{Overview of the BBDA setting for 3D point cloud data.
    A pretrained source model acts as a black-box predictor
    and uses target data for adaptation.}
    \label{fig:3d_sfda}
    \vspace{-1em}
\end{figure}
The self-training generates pseudo-labels of the target and adapts the network with re-training, however, learning with pseudo-labels leads to error accumulation which degrades the performance of the model. To handle this issue, some of the existing work leverages back-propagation free approach for getting the adaptive prediction at the time of inference \cite{wang2024backpropagation,zhang2025backpropagation}. However, the use of source data for the alignment makes it non-feasible. 

The BBDA setting for 3D point cloud data remains largely unexplored within domain adaptation research, illustrated in Fig. \ref{fig:3d_sfda}. 
We study 3D point-cloud adaptation under the joint constraints of
source-free, black-box, and backpropagation-free deployment: source
training data and source-model internals are unavailable, the frozen
source model is accessed only through its predictions, and no
gradient-based parameter updates are performed at test time.
Unlike conventional domain adaptation approaches that rely on access to both source data and the corresponding model, the black-box paradigm restricts access solely to the pre-trained source model, which is treated as an immutable black box(Table \ref{tab:constraints_bftt}).

In this work, we address a stricter and practically motivated setting. We propose a novel \emph{\textbf{Backpropagation-free} Black-box Domain Adaptation framework for 3D point cloud data} (\textbf{3D-BBDA}), that performs adaptation without accessing source data or source model internals and without updating any parameters at test-time.
By avoiding parameter updates, the model eliminates the risk of error accumulation that could degrade performance over time. The source model is treated purely as a black-box predictor. The goal is to reduce domain gap between source and target domains. 

A prototype-based alignment module is leveraged to align the feature space of the source and target domains. A non-parametric network is used to extract features from the target samples. For each corruption, the features from the target samples are extracted using a non-parametric network, and they are assigned labels based on the pseudo labels obtained from the pre-trained source model through the assumption that the pseudo labels are the ground truth. Based on these pseudo-labels, the prototype for each class is generated by taking the mean features corresponding to each label. Further, these class-wise prototypes are used to align the samples. To avoid losing information from the source domain, a weighted prediction module is used to obtain the final prediction by integrating the source model and the non-parametric network.

To ensure stability under sparse or incomplete pseudo-label assignments, we further construct a \emph{Gaussian-perturbed prototype prior} that provides fallback class prototypes when target pseudo-label coverage is missing or unreliable. Finally, a weighted prediction module fuses source predictions with alignment-based predictions, balancing source generalization with target-specific correction while avoiding error accumulation. The resulting framework requires no source data, no source-model parameter access, and no gradient updates, making it suitable for privacy-preserving and deployment-constrained 3D recognition.
The main contributions are summarized as:
\begin{itemize}
\item We formulate and address backpropagation-free black-box domain adaptation for 3D point cloud recognition, under strict source-free and parameter-free constraints motivated by robotic deployment scenarios.

\item We propose a pseudo-label guided prototype-based alignment module for adapting target features without accessing source data or source model internals.

\item We introduce a \emph{Gaussian-perturbed prototype prior} that stabilizes prototype memory when pseudo-label coverage is incomplete under severe target shifts.

\item We conduct extensive experiments across multiple corrupted 3D benchmarks, demonstrating consistent robustness gains with negligible inference overhead.

\end{itemize}

\section{Related Work}

Deep architectures such as PointNet \cite{qi2017pointnet}, and GDANet \cite{xu2021learning}, as well as recent transformer-based models (DAPoinTr \cite{li2024dapointr}, and P3Former \cite{xiao2024position}), achieve strong in-domain performance but can be brittle under distribution shifts (e.g., sensor noise, occlusion, density changes, corruption), motivating adaptation for robust deployment.

Unsupervised domain adaptation (UDA) aligns labeled source and unlabeled target domains, leveraging geometric priors and self-training, e.g., SRDAN~\cite{zhang2021srdan}, graph-based self-distillation~\cite{cardace2023self}, and FAST3D~\cite{fruhwirth2021fast3d}. However, UDA typically requires source data and often access to model internals, which may be infeasible under privacy, licensing constraints, or data portability constraints. Source-free domain adaptation (SFDA) relaxes the source-data requirement but commonly still relies on access to source-model components and pseudo-labeling, which can suffer from confirmation bias and error accumulation.

BBDA further restricts the setting by allowing only black-box predictions from the source model while prohibiting source data. Prior BBDA/SFDA approaches study augmentation and self-learning with uncertainty-aware filtering~\cite{li2024comprehensive,hegde2023source} and teacher/prototype mechanisms to mitigate noisy labels and forgetting~\cite{duan2024source,hegde2024attentive,li2024graph}. Test-time adaptation (TTA) adapts models using unlabeled test data; in 3D, recent work explores reconstruction/augmentation and prototype-based strategies~\cite{wang2020tent,vu2024test,jiang2024pcotta,zou2025hgl}. Closest to our deployment constraints, BFTT3D~\cite{wang2024backpropagation} performs backpropagation-free 3D test-time adaptation via prototype memory for feature alignment, but it uses source data to build the prototype memory. It has conflicts with strict source-free and privacy-constrained deployment.

In contrast, we address a stricter and practically motivated regime: \emph{backpropagation-free black-box adaptation for 3D point clouds}, requiring neither source data nor source-model internals (or gradient updates), while improving robustness under corrupted target conditions, and architecture-agnostic adaptation strategies.
Unlike BFTT3D, which constructs its adaptation memory using source
information, 3D-BBDA replaces source-derived prototypes with
pseudo-label-guided target prototypes, completes missing classes using
an unlabeled target prior, and fuses the resulting prediction with
black-box source output.

\begin{table}[t]
\centering
\scalebox{0.5}{
\begin{tabular}{lccc}
\toprule
\textbf{Setting} & \textbf{Source data} &
\textbf{Model internals} & \textbf{Target BP} \\
\midrule
Source-free      & No & Allowed & Allowed \\
Black-box        & No & No      & Allowed \\
Backprop.-free   & Allowed & Allowed & No \\
\textbf{3D-BBDA} & \textbf{No} & \textbf{No} & \textbf{No} \\
\bottomrule
\vspace{0.01em}
\end{tabular}
}

\scalebox{0.5}{
\begin{tabular}{lcc}
\toprule
\textbf{Property} & \textbf{BFTT3D} & \textbf{3D-BBDA} \\
\midrule
Source data during adaptation & Required & No \\
Source-model internals & Required & No \\
Source-model gradients & No & No \\
Prototype source & Source features & Target pseudo-labels \\
Missing-class completion & No & Gaussian target prior \\
Source/adaptation fusion & No & Yes \\
Black-box source interface & No & Yes \\
\bottomrule
\end{tabular}
}
\vspace{-0.4em}
\caption{Top: distinction between source-free, black-box, and
backpropagation-free adaptation constraints. ``Allowed'' means the
setting alone does not prohibit that access. Bottom: key differences
between BFTT3D and 3D-BBDA.}
\label{tab:constraints_bftt}
\vspace{-1.4em}
\end{table}

\section{Methodology}

\subsection{Problem Statement}
\begin{figure}[t]
\begin{center}
\includegraphics[width=0.9\linewidth]{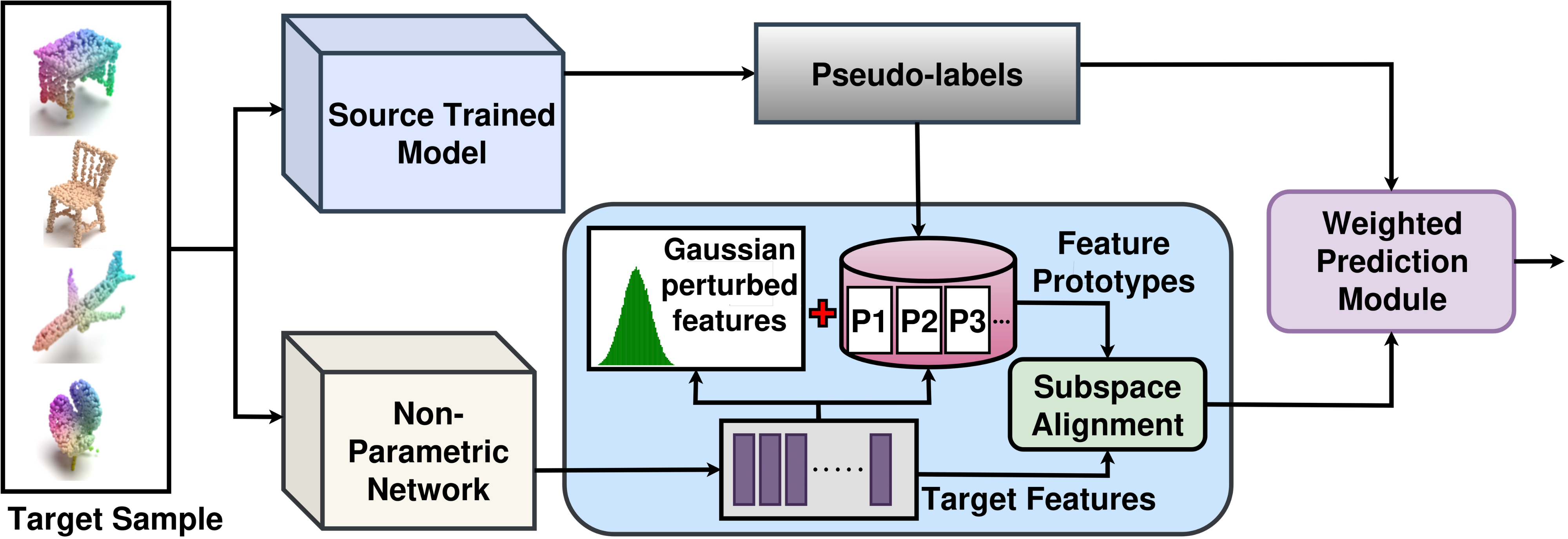}
\end{center}
\vspace{-1.4em}
    \caption{Overview of 3D-BBDA, where target features from a non-parametric network are aligned via prototype generation and shared-subspace alignment. A Gaussian-perturbed prototype prior completes missing class prototypes, and final predictions fuse source and alignment outputs via weighted prediction.}
\label{fig:main_network}
\vspace{-1.5em}
\end{figure}

We address the Black-box domain adaptation problem in 3D point cloud data for the classification task using source trained model as the pre-trained black-box predictor ($F_{S}$), i.e., only prediction outputs are accessible, trained with labeled source domain data $\lbrace x_i^s, y_i^s\rbrace _{i=1}^{n_s}$ under the domain $D_s$, where $n_{s}$ represents the number of source samples (Fig. \ref{fig:main_network}). For adaptation, we have unlabeled target domain data $\lbrace x_i^t\rbrace _{i=1}^{n_t}$ under domain $D_{t}$, where $n_{t}$ represents number of target samples. Here, we make use of a non-parametric network ($F_{nn}$) following the convention of~\cite{wang2024backpropagation}. The objective of the proposed 3D-BBDA framework is to adapt the source model ($F_{S}$) to the target domain at the test time and infer target labels $\lbrace y_i^t\rbrace _{i=1}^{n_t}$ using pre-trained source model, a non-parametric network($F_{nn}$) and unlabeled target data $\lbrace x_i^t\rbrace _{i=1}^{n_t}$ in the absence of source data or internal parameters, ensuring robust perception across unseen conditions. It is summarized in Algorithm in supplementary material.

\noindent \textbf{Closed-set assumption.}
We consider the closed-set adaptation setting, where the source and
target domains share the same label space,
$\mathcal{Y}_s=\mathcal{Y}_t=\{1,\ldots,C\}$.
Although target labels are unavailable during adaptation, each target
sample is assumed to belong to one of the $C$ source classes. This
assumption enables class-wise prototype construction from black-box
pseudo-labels and the subsequent fusion with source-model predictions.

\subsection{Feature Extraction From Non-parametric Network}
For clarity, $x_i^t$ denotes the $i$-th target point-cloud sample
in $\mathcal{D}_t$, whereas $x_{t,i}$ in the following feature
extraction equations denotes an individual 3D point within the
corresponding target point cloud.
A non-parametric network, PointNN~\cite{zhang2023parameter}, is utilized
to extract robust features from the target domain. 
Trigonometric functions first encode its $x$, $y$, and $z$ coordinates
into channel embeddings of size $d$.
For each channel \(s \in \{x, y, z\}\), the channel embeddingas:
\begin{equation}
\scalebox{0.9}{$
\label{trigonometric_equation}
g(x_{t,i})^{(s)} = 
\begin{cases} 
\sin(\alpha x_{t,i}^{(s)} / \beta^{\frac{i}{d}}), & i = 2k \\
\cos(\alpha x_{t,i}^{(s)} / \beta^{\frac{i}{d}}), & i = 2k + 1
\end{cases}
\quad \in \mathbb{R}^{d/3}
$}
\end{equation}
Here, \(\alpha\) and \(\beta\) represent wavelength and scale hyperparameters. 
These channel-wise embeddings from \(x,y,z\) coordinates are concatenated to obtain raw embedding \(f_{\text{raw}, t, i}\) as:
\begin{equation}
    \scalebox{0.9}{$
    \label{embedding}
    f_{\text{raw}, t, i} = \big[ g(x_{t,i})^{(X)}, \, g(x_{t,i})^{(Y)}, \, g(x_{t,i})^{(Z)} \big] \quad \in \mathbb{R}^d
    $}
\end{equation}
Further, Farthest Point Sampling (FPS) is employed to downsample the local center points \(x_{\text{cen}}\) from all points in the dataset. Following this, \(k\)-Nearest Neighbors (\(k\)-NN) is used to group \(N_{\text{ct}}\) neighboring points (\(x_{\text{nb}}\)) for each center point \(x_{\text{cen}}\).
 Using a combination of both center and neighbor information,
 the expanded center feature \(\tilde{f}_{\text{cen}}\) for \(x_{\text{cen}}\) as:
 % is calculated as:
\begin{equation}
\label{fps}
\tilde{f}_{\text{cen}} = \text{Concat}(f_{\text{cen}}, f_{\text{nb},j}), \, j \in N_{\text{cen}}
\end{equation}

Here, \(f_{\text{cen}}\) is feature of point \(x_{\text{cen}}\), \(f_{\text{nb}}\) denotes the feature of point \(x_{\text{nb}}\), and \(\mathcal{N}_{\text{cen}}\) refers to the set of neighboring points for center point \(x_{\text{cen}}\). 
To show the spatial distribution of neighboring points \(x_{\text{nb}}\), feature \(\tilde{f}_{\text{cen}}\) is further reweighted by its relative position encoding \(g(\Delta_{\text{nb}})\) to the center point as
\vspace{-0.2em}
\begin{equation}
    \scalebox{0.9}{$
    \label{positional encoding}
    \tilde{f}_{\text{cen}}^{w} = (\tilde{f}_{\text{cen}} + g(\Delta_{\text{nb}})) \odot g(\Delta_{\text{nb}})
    $}
    \vspace{-0.2em}
\end{equation}
where \(\odot\) represents element-wise multiplication and \(\Delta_{\text{nb}}\) is the relative position of \(x_{\text{nb}}\) to its center point \(x_{\text{cen}}\). Finally, the max pooling and average pooling are used to reduce the dimensionality of the information of \(\tilde{f}_{\text{cen}}^{w}\) as:
\begin{equation}
\vspace{-0.2em}
\scalebox{0.9}{$
\label{maxpooling}
\tilde{f}_{\text{cen}}^{p} = \text{MaxPool}(\tilde{f}_{\text{cen}}^{w}) + \text{AvgPool}(\tilde{f}_{\text{cen}}^{w})
$}
\vspace{-0.2em}
\end{equation}
\noindent Finally, a global pooling operation is applied to get the feature \(f_{t,i} \in \mathbb{R}^d\) for \(x_{t,i}\). The extracted features are used as a basis for prototype generation in the subsequent phase of our alignment framework.

\subsection{Prototype-based Alignment Module}
For each input point cloud $x_{t, i}$, the corresponding pseudo-label $ \hat{y}_i^t$ is obtained through $F_{S}$. Simultaneously, through a non-parametric network, target features  $(f_{t,i})$ are extracted. 
The extracted features are assigned their corresponding pseudo labels.
\textbf{Target-buffer protocol.}
The mini-batch size of 24 is used only for data loading and feature
extraction and does not define the prototype-construction set.
Target features and their black-box pseudo-labels are accumulated
over the unlabeled target buffer, and class prototypes are then
constructed from this accumulated buffer. The resulting prototype
memory is kept fixed while shared-subspace alignment is performed
over target blocks of 128 samples, with $W$ recomputed for each
alignment block.
A feature prototype $(P_{i})$ corresponding to each class $i$ is generated by taking the mean features of all the assigned samples to that class as:
\vspace{-0.8em}
% $ P_i = \frac{1}{n_i} \sum_{j=1}^{n_i} f_j$; 
\begin{equation}
\scalebox{0.9}{$
P_i = \frac{1}{n_i} \sum_{j=1}^{n_i} f_j
\label{eq:feature_prototype}
$}
\vspace{-0.8em}
\end{equation}
where $n_{i}$ is the number of samples assigned to class $i$ and $f_{j}$ represents the feature of  $j^{th}$ sample assigned to class $i$. 

\subsubsection{Gaussian-perturbed prototype prior}
Under severe domain shift, pseudo-labeling may yield incomplete class coverage, where some classes receive no assigned target samples. In such cases, the corresponding prototype in Equation~\ref{eq:feature_prototype} cannot be computed, which may destabilize the subsequent alignment step. To maintain a class-complete prototype memory, we construct a \emph{Gaussian-perturbed prototype prior} from unlabeled target predictions and use it to complete missing class prototypes during alignment.  Specifically, an unlabeled Gaussian-perturbed target stream is passed through the black-box source predictor \(F_S\) and the frozen non-parametric encoder \(F_{nn}\). The resulting pseudo-labels and features are used to compute a class-aligned prototype prior \(\{P_i^{g}\}_{i=1}^{C}\). For class \(i\), let \(\mathcal{S}_i^{g}\) denote the set of Gaussian-perturbed target samples assigned to class \(i\), and let \(n_i^{g}=|\mathcal{S}_i^{g}|\). If a class is missing even in the Gaussian-perturbed stream, we use the global mean of available Gaussian-perturbed prototypes as a conservative fallback:
\vspace{-0.2em}
\begin{equation}
\scalebox{0.9}{$
P_i^{g} =
\begin{cases}
\frac{1}{n_i^{g}}\sum_{k \in \mathcal{S}_i^{g}} f_{t,k}^{g}, & \text{if } n_i^{g} > 0, \\
\frac{1}{|\mathcal{C}_g^{+}|}\sum_{j \in \mathcal{C}_g^{+}} P_j^{g}, & \text{if } n_i^{g} = 0,
\end{cases}
\label{eq:gaussian_bank}
$}
\vspace{-0.2em}
\end{equation}
where \(\mathcal{C}_g^{+}=\{j \mid n_j^{g}>0\}\) is the set of classes with available Gaussian-perturbed prototypes. This prior is computed from unlabeled target predictions and does not require source samples, target labels, source-model parameters, or gradient updates.

\textbf{Prototype completion.}
For a target stream \(d\), pseudo-labeling may leave some classes with no assigned samples. Let \(\mathcal{S}_i^{(d)}\) denote the set of target samples in stream \(d\) assigned to class \(i\), and let \(n_i^{(d)}=|\mathcal{S}_i^{(d)}|\). The completed prototype for class \(i\) is computed as
\vspace{-0.2em}
\begin{equation}
\scalebox{0.9}{$
\tilde{P}_i^{(d)} =
\begin{cases}
\frac{1}{n_i^{(d)}}\sum_{k \in \mathcal{S}_i^{(d)}} f_{t,k}^{(d)}, & \text{if } n_i^{(d)} > 0, \\
P_i^{g}, & \text{if } n_i^{(d)} = 0.
\end{cases}
\label{eq:prototype_fill}
$}
\vspace{-0.2em}
\end{equation}
The completed prototype memory \(\{\tilde{P}_i^{(d)}\}_{i=1}^{C}\) is then used for shared-subspace alignment. This completion step provides fallback class prototypes for missing pseudo-labeled classes while preserving the backpropagation-free, black-box, and source-free constraints of 3D-BBDA.

\begin{table*}[t]
    % \centering
\begin{center}
     \begin{adjustbox}{width=0.8\textwidth}
    \begin{tabular}{c|c|ccccccccccccccc|c}
        \toprule
        Method  & Backbone  &  unif  & gaus  & backg  & impul  & upsamp  & rbf  & rbf-inv  & den-dec  & dens-inc  & shear  &  rot  &  cut  &  dist  &  oclu  &  lidar  & Mean $\downarrow$ \\
        \midrule
        % \multirow{7}{*}{PointNet}
        Source-only & \multirow{10}{*}{PointNet} &  14.63 & 18.68 & 95.30 & 33.39 & 15.03 & 29.46 & 27.63 & 12.93 & 10.49 & 42.71 & 72.81 & 14.95 & 34.85 & 56.28 & 59.00 & 35.88   \\
         TENT \cite{wang2020tent} & & 14.71 & 18.35 & 90.36 & 27.03 & 15.36 & 28.28 & 26.66 & 13.86 & 12.36 & 40.68 & 65.68 & 15.76 & 33.87 & 56.56 & 59.85 & 34.62\\
        BN \cite{li2016revisiting} & & 14.83 & 18.84 & 89.63 & 27.47 & 15.15 & 28.69 & 26.66 & 14.59 & 12.40 & 40.44 & 65.88 & 16.25 & 34.68 & 57.58 & 60.21 & 34.89 \\
        SHOT \cite{liang2020we} & & 14.91 & 17.22 & 90.44 & 25.61 & 14.51 & 27.07 & 25.85 & 13.78 & 12.20 & 39.95 & 65.80 & 16.21 & 32.98 & 56.93 & 60.29 & 34.25 \\
        LAME \cite{boudiaf2022parameter} & & 76.64 & 84.67 & 95.95 & 95.48 & 67.83 & 93.37 & 91.72 & 59.92 & 34.81 & 96.45 & 97.50 & 68.52 & 92.53 & 95.99 & 95.95 & 83.16 \\
        DSS \cite{wang2024continual} & & 14.20 & 18.71 & 94.13 & 33.12 & 14.33 & 29.31 & 27.55 & 13.31 & 10.92 & 41.42 & 66.31 & 15.45 & 34.10 & 56.31 & 59.33 & 35.23 \\
        BFTT3D \cite{wang2024backpropagation}  &  & 14.55 & 18.27 & 80.75 & 31.93 & 14.83 & 28.97 & 27.19 & 12.76 & 10.49 & 39.71 & 68.56 & 14.67 & 34.04 & 54.13 & 55.88 & 33.78   \\
        PURGE-GATE (PG-SP$^\dagger$) \cite{yazdanpanah2025purge}  &
        & 15.44 & 19.37 & 93.40 & 36.63 & 15.19
        & 30.59 & 27.76 & 13.61 & 10.74 & 42.59
        & 72.45 & 14.91 & 33.83 & 56.52 & 59.68
        & 36.18 \\
        DINE-3D$^\ddagger$ \cite{liang2022dine} &
        & 34.90 & 32.10 & 96.02 & 40.68 & 33.08
        & 43.27 & 42.44 & 30.00 & 28.11 & 50.36
        & 66.42 & 30.38 & 46.23 & 59.01 & 64.99
        & 46.53 \\
        % \midrule
        % \hdashline[0.5pt/2pt]  % (dash length / gap length)
        \textbf{\makecell{3D-BBDA (Ours)}}  &   &  15.36 & \textbf{16.95} & 93.63 & \textbf{25.49} & 14.58 & \textbf{24.67} & \textbf{23.08} & \textbf{12.70} & 10.58 & \textbf{37.50} & \textbf{59.56} & \textbf{13.60} & \textbf{28.68} & \textbf{51.84} & \textbf{53.02} & \textbf{32.08} \\
        \midrule
         Source-only & \multirow{5}{*}{GDANet} & 21.73 & 27.17 & 60.95 & 39.99 & 30.27 & 23.28 & 21.81 & 24.47 & 18.67 & 17.89 & 43.67 & 21.12 & 20.96 & 64.83 & 72.83 & 34.03    \\
        BFTT3D \cite{wang2024backpropagation}  &  &  21.08 & 28.02 & 57.76 & 36.44 & 27.04 & 21.98 & 21.49 & 21.57 & 17.24 & 16.75 & 41.87 & 20.14 & 19.49 & 63.32 & 72.06 & 32.36   \\
        PURGE-GATE (PG-SP$^\dagger$) \cite{yazdanpanah2025purge} &
        & 23.18 & 29.78 & 31.85 & 37.80 & 31.48
        & 24.39 & 22.65 & 24.35 & 19.77 & 17.99
        & 44.25 & 22.00 & 20.79 & 65.32 & 73.58
        & 32.61 \\
        DINE-3D$^\ddagger$ \cite{liang2022dine} &
        & 39.01 & 38.13 & 63.30 & 43.94 & 43.33
        & 44.04 & 42.88 & 38.38 & 34.67 & 33.70
        & 55.19 & 34.06 & 42.83 & 72.43 & 74.89
        & 46.72 \\
        % \cmidrule(lr{.5em}){1-3} \cmidrule(lr{.5em}){5-7}
        \textbf{\makecell{3D-BBDA (Ours)}}  & &  \textbf{20.47} & \textbf{23.16} & \textbf{48.37} & \textbf{33.17}  & \textbf{26.27} & \textbf{21.53} & \textbf{20.55} & \textbf{19.93} & \textbf{16.91} & \textbf{15.40} & \textbf{34.80} & \textbf{18.14} & 19.61 & \textbf{58.62} & \textbf{67.40} & \textbf{29.62} \\
        \bottomrule
    \end{tabular}
    \end{adjustbox}
\end{center}
\vspace{-1.8em}
        \caption{Performance  under ModelNet-40C corruptions with PointNet and GDANet \textbf{(Trained from Scratch)}.}
    \label{tab:result_pointnet_modelnet}
    \label{tab:result_GDANet_modelnet}
\vspace{-1em}
\end{table*}

\begin{table*}[t]
    % \centering
\begin{center}
     \begin{adjustbox}{width=0.8\textwidth}
    \begin{tabular}{c|c|ccccccccccccccc|c}
        \toprule
        Method  & Backbone  &  unif  & gaus  & backg  & impul  & upsamp  & rbf  & rbf-inv  & den-dec  & dens-inc  & shear  &  rot  &  cut  &  dist  &  oclu  &  lidar  & Mean $\downarrow$ \\
        \midrule
        % \multirow{7}{*}{PointNet}
        Source-only & \multirow{8}{*}{PointNet} &  52.78 & 58.68 & 89.58 & 64.41 & 50.69 & 51.74 & 49.31 & 43.75 & 41.49 & 57.12 & 60.42 & 44.97 & 51.39 & 93.06 & 95.14 & 60.30   \\
        SHOT \cite{liang2020we} & & 57.8 & \textbf{37.7} & 85.0 & 66.2 & 58.4 & 64.1 & 64.7 & \textbf{34.0} & \textbf{33.6} & 63.7 & 67.2 & 34.6 & 59.6 & \textbf{91.0} & 91.6 & 60.6 \\
        DUA \cite{mirza2022norm} &  & 71.9 & 56.7 & 86.1 & 78.3 & 68.8 & 70.9 & 65.2 & 37.1 & 34.4 & 69.9 & 74.0 & 36.7 & 68.8 & 92.4 & 90.8 & 66.8 \\
        TENT \cite{wang2020tent} & & 59.8 & 38.5 & \textbf{83.8} & 64.3 & 59.2 & 64.3 & 62.9 & 34.6 & \textbf{33.6} & 64.1 & 64.9 & \textbf{33.0} & 60.4 & 91.2 & \textbf{89.9} & 60.3 \\
        BFTT3D \cite{wang2024backpropagation}  &  & 53.30 & 57.99 & 89.41 & 65.62 & 50.87 & 51.39 & 48.44 & 42.36 & 41.15 & 56.60 & 59.72 & 43.75 & 51.04 & 92.88 & 95.14 & 59.98   \\
        PURGE-GATE (PG-SP$^\dagger$) \cite{yazdanpanah2025purge}  &
        & 52.67 & 59.55 & 86.40 & 64.20 & 51.46
        & 52.67 & 50.09 & 44.23 & 42.51 & 56.97
        & 61.62 & 45.09 & 51.29 & 93.12 & 95.01
        & 60.46 \\
        DINE-3D$^\ddagger$ \cite{liang2022dine} &
        & 65.17 & 63.34 & 90.30 & 67.53 & 60.81
        & 61.50 & 58.29 & 58.52 & 52.55 & 61.73
        & 70.22 & 57.83 & 64.31 & 95.87 & 96.39
        & 68.29 \\
        \textbf{\makecell{3D-BBDA (Ours)}}  & &  \textbf{52.26} & \textbf{57.47} & \textbf{87.85} & \textbf{63.54} & \textbf{50.35} & 52.43 & \textbf{47.92} & \textbf{42.36} & \textbf{39.76} & \textbf{56.42} & \textbf{59.38} & 43.92 & \textbf{50.87} & \textbf{92.88} & \textbf{94.79} & \textbf{59.48} \\
        \midrule
        Source-only & \multirow{5}{*}{GDANet} &   71.18 & 40.80 & 83.16 & 71.53 & 75.00 & 68.75 & 68.58 & 43.40 & 35.59 & 66.84 & 71.70 & 38.19 & 70.83 & 92.71 & 94.10 & 66.16   \\
        BFTT3D \cite{wang2024backpropagation}  &  &   72.57 & 41.15 & 83.16 & 71.70 & 74.48 & 69.27 & 69.10 & 42.19 & 34.38 & 67.36 & 71.18 & 37.15 & 71.18 & 92.88 & 94.10 & 66.12  \\
        PURGE-GATE (PG-SP$^\dagger$) \cite{yazdanpanah2025purge} &
        & 71.08 & 40.28 & 79.35 & 70.05 & 75.04
        & 69.19 & 68.67 & 43.72 & 35.80 & 66.78
        & 71.43 & 37.52 & 70.74 & 92.43 & 93.63
        & 65.71 \\
        DINE-3D$^\ddagger$ \cite{liang2022dine} &
        & 75.90 & 52.15 & 81.30 & 75.96 & 74.64
        & 73.95 & 77.45 & 55.54 & 51.46 & 78.43
        & 80.32 & 51.75 & 78.08 & 88.18 & 88.93
        & 72.27 \\
        \textbf{\makecell{3D-BBDA (Ours)}}  & &  71.35 & \textbf{39.58} & \textbf{82.29} & \textbf{70.66} & 75.35 & 69.44 & \textbf{68.06} & 42.36 & \textbf{34.20} & \textbf{65.97} & 71.70 & 37.85 & \textbf{69.79} & \textbf{92.19} & \textbf{93.75} & \textbf{65.64 } \\
        \bottomrule
    \end{tabular}
    \end{adjustbox}
\end{center}
\vspace{-1.8em}
    \caption{Performance  under ScanObjectNN-C corruptions with PointNet and GDANet \textbf{(Trained from Scratch)}.}
    \label{tab:result_ScanObjectNN_modelnet}
\vspace{-1.7em}
\end{table*}

% \vspace{-2em}
\subsubsection{Reducing domain shift through shared subspace projection and alignment} 
Domain shifts occur when target features diverge from the distribution of learned prototypes, rendering direct alignment inadequate.
It begins by computing cosine similarity normalized feature vectors from the target domain and prototype memory to guide classification \cite{wang2024backpropagation}. Consider a target feature vector  \( G_t = [g_{t,1}, g_{t,2}, \dots, g_{t,n_t}] \) and prototype memory \( P = [P_1, P_2, \dots, P_c] \). The cosine similarity as:
\vspace{-0.3em}
\begin{equation}
\scalebox{1}{$
\label{similarity matrix}
K = \frac{\widetilde{G}_t \cdot \widetilde{P}}{\|\widetilde{G}_t\| \|\widetilde{P}\|}
$}
\vspace{-0.3em}
\end{equation}
Here, vector \(\widetilde G_t \) represents the normalized feature vector of \( G_t \), expressed as \( \widetilde{G}_t = \frac{G_t}{\|G_t\|} \), where \( \|G_t\| \) denotes the magnitude of \( G_t \). Similarly, \(\widetilde P \) represents the normalized prototype memory vector, given by \( \widetilde{P} = \frac{P}{\|P\|} \), where \( \|P\| \) represents the magnitude of \( P \). This similarity leads to classification by assigning a weighting to the influence of each prototype. The output logits with pseudo label \( y_i \) as:
\begin{equation}
\vspace{-0.4em}
\scalebox{0.9}{$
\label{similarity logits}
l_{am} = \phi(K_{y_i}), \quad \phi(x) = \exp(-\beta(1-x)),
$}
\vspace{-0.2em}
\end{equation}
with \( \beta \) as a scaling hyperparameter.
To explicitly reduce the distribution gap between prototypes ($f^*$) and target features ($f_t^*$), both are projected into a shared subspace using a mapping function $\psi(.)$. The disparity between these domains is reduced by calculating the \textbf{statistical distance} between their mean representations:
\vspace{-0.2em}
\begin{equation}
\scalebox{0.9}{$
\label{distance_p_g_t}
\text{Dist}(P, G_t) = \left\| \frac{1}{n_p} \sum_{i=1}^{n_p} f_i^* - \frac{1}{n_t} \sum_{i=1}^{n_t} f_t^* \right\|_2^2.
$}
\vspace{-0.4em}
\end{equation}
\noindent where $ n_p$ and $ n_t$ represent the number of prototype features and target features, respectively. 
Now, we introduce a kernel similarity matrix $K$ and a scaling matrix $L$, which encodes similarities between prototypes and test data points as $K_{i,j} = \langle \psi(x_i), \psi(x_j) \rangle$. The value of \(L_{i,j}\) is defined as:
\vspace{-0.4em}
\begin{equation}
\scalebox{0.9}{$
L_{i,j} =
\begin{cases} 
\frac{1}{n_t^2} & \text{if } x_i, x_j \in G_t, \\
\frac{1}{n_p^2} & \text{if } x_i, x_j \in P, \\
-\frac{1}{n_p n_t} & \text{otherwise.}
\end{cases}
$}
\vspace{-0.4em}
\end{equation}
where \(x_i\) and \(x_j\) are points being compared. The equation assigns values to \(L_{i,j}\) based on whether both points belong to the target domain (\(G_{t}\)), the prototype ($P$), or are from different domains. To further refine the alignment, the transformation matrix $W$ is optimized by minimizing the trace regularization objective:
\vspace{-0.4em}
\begin{equation}
\scalebox{0.8}{$
\min_W \text{tr}(W^T KLKW) + \mu \, \text{tr}(W^T W),
$}
\vspace{-0.4em}
\end{equation}
\noindent
where $\mu$ controls the complexity of
$W \in \mathbb{R}^{(n_p+n_t)\times m}$,
and the centering matrix is
$H = I-\frac{1}{n_p+n_t}\mathbf{1}\mathbf{1}^{T}
\in \mathbb{R}^{(n_p+n_t)\times(n_p+n_t)}$.
The optimal solution of $W$ is obtained from the $m$ largest
eigenvectors of $(KLK+\mu I)^{-1}KHK$.
The resulting transformation projects the target-derived prototypes
($f^*$) and target features ($f_t^*$) in
Eq.~\ref{distance_p_g_t} into the shared latent subspace through
$\mathbf{W}^{T}\mathbf{K}
\in \mathbb{R}^{m\times(n_p+n_t)}$,
thereby reducing domain divergence.
By aligning global statistics and local similarities, the model
improves robustness under domain shift.

\begin{table*}[t]
    % \vspace{-1em}
    % \centering
\begin{center}
     \begin{adjustbox}{width=0.75\linewidth}
    \begin{tabular}{c|c|ccccccccccccc|c}
        \toprule
        Method  & Backbone  &  unif  & gaus  & backg  & impul  & upsamp  & rbf  & rbf-inv  & den-dec  & dens-inc  & shear  &  rot  &  cut  &  dist  & Mean $\downarrow$ \\
        \midrule
        % \multirow{7}{*}{PointNet}
        Source-only & \multirow{5}{*}{PointNet}    & 35.87 & 22.96 & 18.10 & 48.45 & 29.11 & 43.64 & 42.91 & 20.02 & 17.27 & 49.20 & 66.46 & 21.96 & 50.29  &  35.86  \\
        BFTT3D \cite{wang2024backpropagation}  &   & 34.76 & 23.80 & 17.78 & 43.64 & 28.44 & 40.54 & 39.56 & 19.41 & 17.20 & 44.40 & 57.61 & 21.13 & 45.90  &  33.40  \\
        PURGE-GATE (PG-SP$^\dagger$) \cite{yazdanpanah2025purge}  & 
        & 37.19 & 24.33 & 18.46 & 50.38 & 29.82
        & 44.32 & 43.61 & 20.51 & 17.53 & 49.64
        & 66.92 & 22.32 & 50.38
        & 36.57 \\
        DINE-3D$^\ddagger$ \cite{liang2022dine} &
        & 59.75 & 59.73 & 57.67 & 65.22 & 59.76
        & 62.20 & 63.29 & 58.39 & 57.65 & 62.95
        & 64.70 & 57.36 & 63.09
        & 60.90 \\
        \textbf{\makecell{3D-BBDA (Ours)}}   &   &  \textbf{33.29} & 23.27 & 18.08 & \textbf{43.12} & \textbf{28.14} & \textbf{37.05} & \textbf{36.23} & 19.53 & 17.84 & \textbf{43.17} & 61.46 & \textbf{20.91} & \textbf{43.38} &  \textbf{32.74} \\
        \midrule
        Source-only & \multirow{5}{*}{GDANet}    &  29.35 & 26.14 & 13.61 & 37.82 & 15.57 & 24.17 & 24.09 & 21.29 & 16.30 & 21.68 & 33.15 & 22.32 & 23.17 &   23.74  \\
        BFTT3D \cite{wang2024backpropagation}   &    & 44.86 & 47.38 & 28.12 & 47.43 & 28.09 & 44.38 & 43.88 & 32.77 & 28.99 & 40.98 & 56.76 & 33.17 & 44.71  &  40.12   \\
        PURGE-GATE (PG-SP$^\dagger$) \cite{yazdanpanah2025purge}  & 
        & 30.67 & 27.23 & 14.13 & 36.25 & 15.79
        & 25.02 & 25.10 & 21.93 & 17.30 & 21.93
        & 34.19 & 23.13 & 24.24
        & 24.38 \\
        DINE-3D$^\ddagger$ \cite{liang2022dine} &
        & 55.11 & 55.48 & 54.99 & 53.03 & 55.27
        & 56.23 & 54.43 & 53.63 & 54.54 & 55.02
        & 54.92 & 52.62 & 56.37
        & 54.74 \\
        \textbf{\makecell{3D-BBDA (Ours)}}   &  &  \textbf{27.45} & \textbf{24.90} & 13.82 & \textbf{35.09} & 15.81 & \textbf{23.70} & \textbf{23.37} & \textbf{20.54} & \textbf{16.25} & \textbf{20.76} & \textbf{31.06} & \textbf{21.65} & \textbf{22.72} &  \textbf{22.85} \\
        \bottomrule
    \end{tabular}
    \end{adjustbox}
\end{center}
\vspace{-1.8em}
    \caption{Performance under ShapeNet-C corruptions with PointNet and GDANet \textbf{(Trained from Scratch)}.}
    \label{tab:result_ShapeNet_modelnet}
\vspace{-0.8em}
\end{table*}

\begin{table*}[t]

\begin{center}
     \begin{adjustbox}{width=0.7\linewidth}
    \begin{tabular}{c|ccccccccccccc|c}
        \toprule
        Method    &  unif  & gaus  & backg  & impul  & upsamp  & rbf  & rbf-inv  & den-dec  & dens-inc  & shear  &  rot  &  cut  &  dist  & Mean $\downarrow$ \\
        \midrule

        Source-only     & 89.45 & 57.20 & 85.85 & 89.81 & 88.79 & 89.76 & 89.71 & 31.48 & 31.28 & 89.51 & 89.76 & 31.58 & 90.02 & 73.40  \\
        
        BFTT3D \cite{wang2024backpropagation}     & 89.45 & 57.20 & 85.80 & 89.81 & 88.79 & 89.76 & 89.61 & 31.43 & 31.17 & 89.51 & 89.66 & 31.48 & 89.66 & 73.33  \\
        PURGE-GATE (PG-SP$^\dagger$) \cite{yazdanpanah2025purge} 
        & 89.11 & 57.21 & 85.84 & 89.47 & 88.55
        & 89.37 & 89.52 & 31.34 & 31.24 & 89.31
        & 89.26 & 31.60 & 89.67
        & 73.19 \\
        DINE-3D$^\ddagger$ \cite{liang2022dine}
        & 91.21 & 58.33 & 90.05 & 87.03 & 91.79
        & 89.93 & 86.06 & 51.93 & 51.26 & 86.74
        & 85.62 & 51.01 & 90.73
        & 77.82 \\
        \textbf{\makecell{3D-BBDA (Ours)}}     &  \textbf{88.22} & \textbf{55.92} & 85.91 & \textbf{88.17} & \textbf{88.22} & \textbf{88.22} & \textbf{88.12} & \textbf{30.66} & \textbf{30.92} & \textbf{88.84} & \textbf{88.84} & \textbf{31.38} & \textbf{88.01} & \textbf{72.42} \\
        \bottomrule
    \end{tabular}
    \end{adjustbox}
\end{center}
\vspace{-1.8em}
    \caption{Performance under SydneyUrban-C corruptions with PointNet backbone \textbf{(Trained from Scratch)}.}
    \label{tab:result_SydneyUrban_modelnet}
    \vspace{-1.6em}
\end{table*}
\subsection{Weighted Prediction Module}
In order to make our framework robust against the limited number of target samples, using alone the prediction from the alignment module might not work well as it deals with target features only. And we do not want to lose the information of the source model as it will be helpful in cases, where our target sample is close in distribution to the source domain. Additionally, using only the output from the source model might not work well due to the internal covariate shift arising because of the domain gap in source and target distribution.
Thus, in order to incorporate the prediction from both, the source model and the alignment module, we adopt a weighted prediction module as:
\vspace{-0.4em}
\begin{equation}
\scalebox{1}{$
\label{final logit}
l_{\text{final}} = \lambda \cdot l_s + (1 - \lambda) \cdot l_{am}
$}
\vspace{-0.4em}
\end{equation}
Here, $l_{s}$ is the logits obtained from the pre-trained source model, $l_{am}$ is the logits obtained from the alignment module, $l_{final}$ is the final logit which will be used for prediction, and $\lambda$ is the weight assigned to the prediction from the source model.  
We use a predefined $\lambda$ because we wanted to intentionally assign less weight to the source model’s predictions to reduce the impact of noisy labels on the final prediction. 
We treat $\lambda$ as a fixed hyperparameter controlling the
contribution of the source prediction. The originally reported
benchmark-specific values were selected through one-time offline
validation sweeps analogous to Table \ref{tab:lambda_sensitivity} and were subsequently kept
fixed for all samples and corruption types within each benchmark.
This validation uses labels only for offline hyperparameter selection
and is not part of the adaptation pipeline; no target labels are
accessed during pseudo-label generation, prototype construction,
alignment, or inference.
To assess whether 3D-BBDA depends on benchmark-specific selection,
we additionally fix a single global $\lambda=0.1$ and transfer it
unchanged across datasets. This setting still improves Source-only
by +3.67, +1.06, and +2.76 percentage points on ModelNet40-C,
ScanObjectNN-C, and ShapeNet-C, respectively (Table~\ref{tab:global_lambda}).
Table \ref{tab:noisy_pseudo-labels} shows that this strategy makes our framework less sensitive to noisy predictions. 
\begin{table}[t]
\centering

\scalebox{0.6}{
\begin{tabular}{cccccc}
\toprule
$\lambda$ & 0 & 0.1 & 0.2 & 0.5 & 0.8 \\
\midrule
Mean Error (\%) $\downarrow$
& 54.45
& \textbf{32.08}
& 32.74
& 34.85
& 35.57 \\
\bottomrule
\end{tabular}
}
\vspace{-0.8em}
\caption{\footnotesize Sensitivity to fusion weight $\lambda$ on
ModelNet40-C with PointNet.}

\label{tab:lambda_sensitivity}
\vspace{-0.8em}
\end{table}

\begin{table}[t]
\centering

\scalebox{0.6}{
\begin{tabular}{lccc}
\toprule
\textbf{Dataset} &
\textbf{Source-only $\downarrow$} &
\textbf{3D-BBDA $\downarrow$} &
\textbf{Gain $\uparrow$} \\
\midrule
ModelNet40-C   & 35.86 & \textbf{32.19} & +3.67 \\
ScanObjectNN-C & 60.55 & \textbf{59.49} & +1.06 \\
ShapeNet-C     & 35.87 & \textbf{33.11} & +2.76 \\
\midrule
Average        & 44.09 & \textbf{41.59} & +2.49 \\
\bottomrule
\end{tabular}
}
\vspace{-0.8em}
\caption{
Performance with a single global fusion weight $\lambda=0.1$
transferred unchanged across datasets using PointNet.
}
\vspace{-1.8em}
\label{tab:global_lambda}
\end{table}

\section{Experimental Setup}
\subsection{Datasets}

\noindent \textbf{\textit{ScanObjectNN-C}} \cite{wang2023sample}: 15 different corruptions are utilized to build this dataset from 581 samples (object-only test set) to evaluate robustness. 
\noindent \textbf{\textit{ModelNet-40C}} \cite{sun2022benchmarking}:
It is a robust benchmark comprising 15 corruption types applied to the original 2468 test set samples, introducing real-world noise to evaluate our model on real-world challenges.  
\noindent \textbf{\textit{ShapeNet-C}} \cite{dastmalchi2025test}: It is a large-scale point cloud dataset with 51,127 shapes across 55 categories. with 13 types of corruptions to the test set similar to ModelNet-40-C \cite{sun2022benchmarking}. 
\noindent \textbf{\textit{SydneyUrban-C}}:  Sydney Urban Objects Dataset \cite{DeDeuge2013UnsupervisedOutdoor3DScans} contains 631 urban road objects scanned with a Velodyne HDL-64E LIDAR, collected in Sydney, Australia. For experiments, we applied 13 types of corruptions to the test set, similar to ModelNet-40C \cite{sun2022benchmarking}, for 14 classes.  
It provides labeled real object instances and the LiDAR viewpoint.

\begin{table*}[ht]

\begin{center}
     \begin{adjustbox}{width=0.7\textwidth}
    \begin{tabular}{c|c|c|ccccccccccccccc|c}
        \toprule
        Case & setting  & $\lambda$  &  unif  & gaus  & backg  & impul  & upsamp  & rbf  & rbf-inv  & den-dec  & dens-inc  & shear  &  rot  &  cut  &  dist  &  oclu  &  lidar  & Mean $\downarrow$ \\
        \midrule
         Case 1 & \makecell{only \\ Source ($l_{s}$)} & 1 &  14.63 & 18.68 & 95.30 & 33.39 & 15.03 & 29.46 & 27.63 & 12.93 & 10.49 & 42.71 & 72.81 & 14.95 & 34.85 & 56.28 & 59.00 & 35.88  \\
         \midrule
        Case 2 & \makecell{alignment \\ module ($l_{am}$)} & 0 &  46.08 & 43.38 & \textbf{92.48} & 47.79 & 46.57 & 49.71 & 50.20 & 44.20 & 42.77 & 55.15 & 72.75 & 43.87 & 55.23 & 62.66 & 63.93 & 54.45  \\
        \midrule
        \multirow{4}{*}{Case 3} & \multirow{4}{*}{ \makecell{Combination \\ of \\ $l_{s}$ and $l_{am}$}} &  0.1 & 15.36 & 16.95 & 93.63 & \textbf{25.49} & 14.58 & \textbf{24.67} & 23.08 & 12.70 & 10.58 & \textbf{37.50} & \textbf{59.56} & 13.60 & \textbf{28.68} & \textbf{51.84} & \textbf{53.02} & \textbf{32.08}  \\
        
        &  &  0.2 &  \textbf{14.22} & \textbf{16.50} & 94.77 & 26.59 & \textbf{13.68} & 24.75 & \textbf{22.79} & \textbf{12.21} & \textbf{10.29} & 38.32 & 63.73 & \textbf{13.28} & 30.60 & 53.59 & 55.76 & 32.74  \\
        
         &  &  0.5 &  14.38 & 17.65 & 95.06 & 31.37 & 14.46 & 27.74 & 26.43 & 12.58 & 10.62 & 40.73 & 70.87 & 14.67 & 33.05 & 54.98 & 58.17 & 34.85 \\
         
          &  &  0.8 & 14.58 & 18.34 & 95.10 & 32.97 & 14.87 & 28.80 & 27.25 & 12.75 & 10.54 & 42.16 & 72.30 & 14.83 & 34.64 & 55.88 & 58.62 & 35.57 \\
 
        \bottomrule
    \end{tabular}
    \end{adjustbox}
\end{center}
\vspace{-1.8em}
    \caption{Ablation on different $\lambda$ values used in the weighted prediction module on the ModelNet-40C dataset with PointNet backbone.}
    \label{tab:ablation_lambda}
    \vspace{-1em}
\end{table*}

\begin{table*}[ht]
\begin{center}
     \begin{adjustbox}{width=0.7\textwidth}
    \begin{tabular}{c|c|c|c|ccccccccccccccc|c}
        \toprule
        \makecell{Disjoint \\ Sets} & \makecell{Alignment \\ set} & \makecell{Prototype \\ set} & Method    &  unif  & gaus  & backg  & impul  & upsamp  & rbf  & rbf-inv  & den-dec  & dens-inc  & shear  &  rot  &  cut  &  dist  &  oclu  &  lidar  & Mean $\downarrow$ \\
        \midrule
        \multirow{2}{*}{a,b,c,d}
        & a & b,c,d & Source-only & \textbf{14.87} & 19.28 & 95.51 & 32.84 & \textbf{14.95} & 29.49 & 28.92 & 13.24 & 10.54 & 43.30 & 72.63 & 14.38 & 33.50 & 58.17 & 59.40 & 36.07  \\
        & a & b,c,d & \textbf{\makecell{3D-BBDA \\ (Ours)}}   &   15.93 & \textbf{18.22} & \textbf{93.87} & \textbf{25.00} & 15.69 & \textbf{24.67} & \textbf{28.19} & \textbf{12.50} & \textbf{11.52} & \textbf{36.76} & \textbf{72.30} & \textbf{12.91} & \textbf{29.58} & \textbf{54.00} & \textbf{54.66} & \textbf{33.72} \\
        \midrule
        \multirow{2}{*}{a,b,c,d} & a,b & c,d &  Source-only & \textbf{13.64} & 20.51 & 95.51 & 32.03 & \textbf{14.87} & 29.00 & 24.35 & 14.30 & 10.70 & 43.22 & \textbf{73.45} & 15.36 & 34.72 & 55.56 & 58.99 & 35.75    \\
        & a,b & c,d &  \textbf{\makecell{3D-BBDA \\ (Ours)}}  & 15.28 & \textbf{18.55} & \textbf{94.2}8 & \textbf{26.88} & 15.20 & \textbf{25.25} & \textbf{21.24} & \textbf{14.30} & \textbf{10.54} & \textbf{39.05} & 76.06 & \textbf{14.62} & \textbf{30.47} & \textbf{52.37} & \textbf{55.07} & \textbf{33.94}   \\
        \midrule
        \multirow{2}{*}{a,b,c,d} & c,d & a,b &  Source-only & 1\textbf{3.83} & 18.67 & 95.50 & 32.00 & \textbf{12.67} & 31.17 & 31.67 & \textbf{12.33} & 9.83 &\textbf{ 44.17} & \textbf{71.33} & 14.83 & 34.17 & 56.17 & 60.67 & 35.93  \\
      & c,d & a,b &  \textbf{\makecell{3D-BBDA \\ (Ours)}}  &  16.00 & \textbf{17.33} & \textbf{90.33} & \textbf{26.00} & 14.50 & \textbf{28.83} & \textbf{24.83} & 13.00 & \textbf{9.17} & 56.00 & 78.33 & \textbf{14.83} & \textbf{28.83} & \textbf{53.33} & \textbf{54.83} & \textbf{35.08} \\
        \bottomrule
    \end{tabular}
    \end{adjustbox}
\end{center}
\vspace{-1.8em}
        \caption{ Ablation on the use of disjoint sets for prototype creation and alignment sets on ModelNet-40C dataset using PointNet for validation of our method. Here, \(a, b, c, \text{and } d\) are disjoint sets, each representing $25\%$ of the target samples.} 
    \label{tab:ablation_disjoint_set}
    \vspace{-1.3em}
\end{table*}

\subsection{Implementation Details}
We use PyTorch library on NVIDIA A100 GPU (80GB). PointNet and GDANet are used as a backbone for source models.  
The batch size is 24, with 1024 points per sample and
corruption severity 5. Target features and black-box pseudo-labels
are accumulated across batches to construct the prototype memory;
shared-subspace alignment is then performed in 128-sample blocks,
with $W$ recomputed for each block.
There are no gradient updates, just a single pass. 
For ModelNet-40C and ScanObjectNN-C datasets, $\lambda$ value is 0.1 and 0.2, respectively. For ShapeNet-C and SydneyUrban-C, $\lambda$ values are 0.05 for PointNet and 0.3 for the GDANet model. For evaluation metrics, we use average classification errors (\%), a decrease in error represents improved performance.
Backbones are trained from scratch (independently trained) for the source model on the corresponding clean training sets to validate robustness of 3D-BBDA.
We utilize pre-trained PointNet backbone only for the ModelNet40 dataset. 
% The pre-trained models are used to adapt to the benchmark corruption datasets, with the 
We use same hyperparameters as in the baseline \cite{wang2024backpropagation}. 
The point-NN model is used for a non-parametric network.
 While our evaluation focuses on established corruption benchmarks, these perturbations emulate common sensing and environmental effects encountered in robotics (e.g., noise, occlusion, density, and motion-induced distortions). 

Native PURGE-GATE PG-SF operates on Transformer-specific
CLS/Q/K representations and is therefore not directly applicable to
PointNet/GDANet. We implement a controlled PG-SP port  by applying
purging to the earliest 64-D point-wise representation using
source-derived embedding statistics with $^\dagger$ as shown in Table \ref{tab:result_pointnet_modelnet}-\ref{tab:result_SydneyUrban_modelnet}.

DINE-3D$^\ddagger$ denotes a controlled feature-space port of
DINE~\cite{liang2022dine}, originally proposed for 2D black-box
domain adaptation. We use same frozen PointNN target features as
3D-BBDA while retaining DINE's source-probability distillation,
EMA memory, information-maximization, MixUp, and target fine-tuning
objectives. Unlike 3D-BBDA, DINE-3D requires target-side
backpropagation.

\section{Results \& Discussion }

\noindent \textbf{ModelNet-40C}

Table~\ref{tab:result_pointnet_modelnet} reports corruption robustness on ModelNet-40C. With PointNet, \textbf{3D-BBDA} achieves lowest mean error (\textbf{32.08\%}), improving over Source-only (35.88\%) and BFTT3D (33.78\%). Gains are particularly pronounced under challenging geometric and sensing shifts, including \emph{rotation} (59.56 vs.\ 68.56), \emph{distortion} (28.68 vs.\ 34.04), and \emph{impulse} (25.49 vs.\ 31.93). With GDANet, \textbf{3D-BBDA} further reduces mean error to \textbf{29.62\%}, outperforming Source-only (34.03\%) and BFTT3D (32.36\%). The strongest improvement occur under difficult corruptions such as \emph{occlusion} (58.62 vs.\ 63.32) and \emph{lidar} (67.40 vs.\ 72.06), showing that 3D-BBDA is effective across backbones.
\noindent \textbf{ScanObjectNN-C}
Table~\ref{tab:result_ScanObjectNN_modelnet} summarizes results on ScanObjectNN-C, which exhibits stronger real-scan characteristics and larger shifts. Despite the increased difficulty, \textbf{3D-BBDA} achieves the best mean error from 60.30\% to \textbf{59.48\%} on PointNet and from 66.16\% to \textbf{65.64\%} on GDANet. Improvements are consistent across many corruptions (e.g., \emph{uniform}: $-1.04\%$ on PointNet and $-1.22\%$ on GDANet vs.\ BFTT3D), indicating stable adaptation even when pseudo-label noise is higher. 
We note that several UDA/SFDA baselines in Table~\ref{tab:result_ScanObjectNN_modelnet} allow source access and/or gradient-based updates, which are outside our strict black-box, source-free, and backpropagation-free setting, prioritizing privacy-preserving adaptation and architectural flexibility.
\noindent \textbf{ShapeNet-C}
Table~\ref{tab:result_ShapeNet_modelnet} shows that \textbf{3D-BBDA} generalizes to the larger-scale ShapeNet-C benchmark. With PointNet trained from scratch, 3D-BBDA reduces mean error from 35.86\% (Source-only) to \textbf{32.74\%} and improves over BFTT3D (33.40\%), with notable gains under RBF/RBF-inv, shear, and distance shifts. With GDANet, 3D-BBDA achieves the lowest mean error (\textbf{22.85\%}) compared to Source-only (23.74\%), while BFTT3D degrades substantially (40.12\%), suggesting that source-dependent prototype construction can be brittle when transferring across datasets and backbones.
These results indicate that 3D-BBDA delivers more stable, backbone-robust adaptation under strict source-free constraints, while remaining competitive across a wide range of realistic corruption patterns.

\begin{figure}[b]
\vspace{-1em}
\begin{center}
    \includegraphics[width=0.75\linewidth] {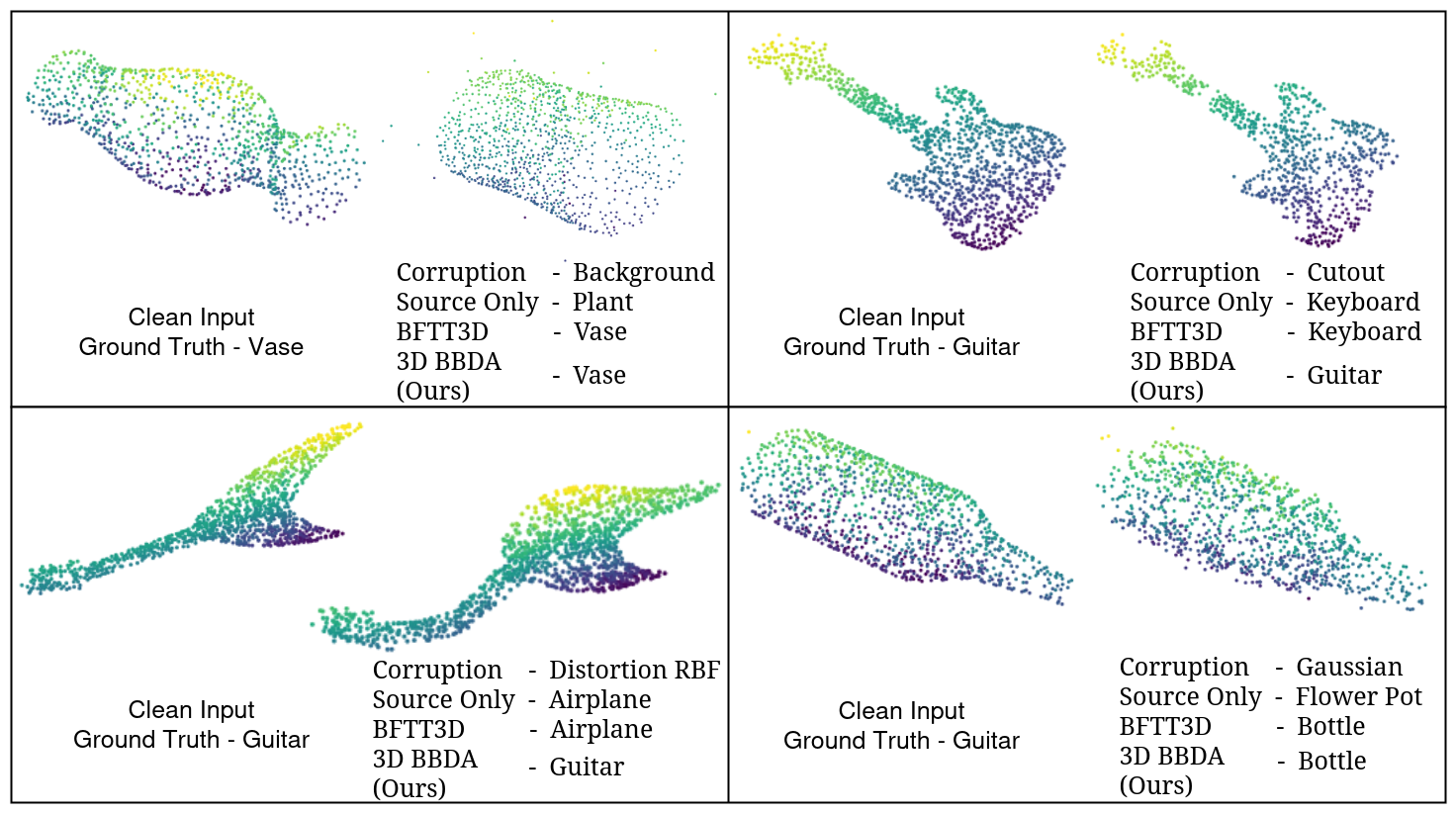}
\end{center}
    \vspace{-1.5em}
    \caption{Qualitative comparison under common corruptions on ModelNet-40C. 3D-BBDA better preserves correct predictions than Source-only and BFTT3D \cite{wang2024backpropagation}.}
    \label{fig:visual_result}
    \vspace{-1em}
\end{figure}

\begin{table*}[t]
    \begin{center}
     \scalebox{0.64}{
    \begin{tabular}{c|ccccccccccccccc|c}
        \toprule
        Method    &  unif  & gaus  & backg  & impul  & upsamp  & rbf  & rbf-inv  & den-dec  & dens-inc  & shear  &  rot  &  cut  &  dist  &  oclu  &  lidar  & Mean $\downarrow$ \\
        \midrule
        BFTT3D \cite{wang2024backpropagation} & 0.584 & 0.539 & \textbf{0.083} & 0.524 & 0.633 & 0.529 & 0.541 & 0.628 & 0.642 & 0.492 & 0.297 & 0.594 & 0.521 & 0.572 & 0.742 & 0.528 \\
        \textbf{ \makecell{3D-BBDA (Ours)}} & \textbf{0.442} &\textbf{ 0.421} & 0.142 & \textbf{0.448} & \textbf{0.486} & \textbf{0.409} & \textbf{0.425} & \textbf{0.575} & \textbf{0.584} & \textbf{0.311} & \textbf{0.049} & \textbf{0.572} & \textbf{0.407} & \textbf{0.370} & \textbf{0.378} & \textbf{0.401} \\
        \bottomrule
    \end{tabular}    }
    \end{center}    
\vspace{-1.8em}
         \caption{\footnotesize Comparison of ECE between BFTT3D and 3D-BBDA on ModelNet-40C.}
        \vspace{-1.2em}
    \label{tab:result_ECE}
\end{table*}

\begin{table}[t]

\begin{center}
    \scalebox{0.34}{
    \begin{tabular}{c|ccccccccccccccc|c}
    \toprule
        Noise    &  unif  & gaus  & backg  & impul  & upsamp  & rbf  & rbf-inv  & den-dec  & dens-inc  & shear  &  rot  &  cut  &  dist  &  oclu  &  lidar  & Mean $\downarrow$ \\
        \midrule
        5\% & 15.36 & 16.87 & 93.83 & 26.10 & 14.62 & 24.31 & 23.41 & 12.62 & 10.29 & 37.21 & 63.36 & 13.73 & 29.21 & 52.78 & 52.74 & 32.43 \\
        20\% & 15.24 & 17.08 & 94.28 & 27.37 & 14.46 & 25.29 & 24.35 & 12.38 & 10.21 & 38.64 & 64.58 & 13.36 & 30.02 & 52.29 & 54.00 & 32.90 \\
        50\% & 14.91 & 17.36 & 94.93 & 29.45 & 14.05 & 26.23 & 24.35 & 12.05 & 10.74 & 40.85 & 67.69 & 13.03 & 31.62 & 53.27 & 56.29 & 33.79 \\
        80\% & 14.34 & 17.69 & 95.10 & 31.99 & 14.22 & 28.15 & 26.10 & 12.54 & 10.29 & 42.65 & 72.02 & 13.85 & 33.46 & 55.11 & 58.70 & 35.08 \\
        100\% & 14.87 & 17.97 & 95.02 & 34.60 & 15.07 & 30.51 & 27.98 & 13.48 & 10.91 & 44.85 & 73.86 & 15.69 & 36.56 & 57.27 & 59.56 & 36.55 \\
        \bottomrule
    \end{tabular}
    }
    
\end{center}
\vspace{-1.8em}
        \caption{ Quality of the pseudo-labels over proposed 3D-BBDA.}
    \label{tab:noisy_pseudo-labels}
\end{table}

\noindent \textbf{SydneyUrban-C}
Table~\ref{tab:result_SydneyUrban_modelnet} evaluates robustness on \emph{SydneyUrban-C}, a real-world LiDAR object dataset corrupted using the same protocol as ModelNet-40C \cite{sun2022benchmarking}. Although absolute errors are high due to the stronger domain gap,
\textbf{3D-BBDA} consistently improves robustness over both Source-only and BFTT3D, reducing the mean error from 73.40\% to \textbf{72.42\%}. 
These results provide evidence that the proposed black-box, backpropagation-free adaptation remains beneficial when moving from synthetic benchmarks to real LiDAR point clouds.

\noindent \textbf{Qualitative comparison under various corruptions }
Fig. \ref{fig:visual_result} presents qualitative comparisons of point cloud classification performance under various corruption types, including background, cutout, distortion rbf, and Gaussian noise. While the baseline approaches (Source Only and BFTT3D) frequently misclassify corrupted samples, our proposed 3D-BBDA consistently preserves the correct semantic labels. These results highlight that 3D-BBDA effectively handles distributional shifts and maintains classification reliability 

\section{Ablation Analysis}

\noindent \textbf{Analysis of Module Contributions Through $\lambda$}
Table~\ref{tab:ablation_lambda} analyzes the contribution of each component by varying  $\lambda$ (Eq. \ref{final logit}).
\noindent\textbf{Case 1 ($\lambda = 1$):} Using only the source model predictions yields a mean error of \textbf{35.86\%}, corresponding to the source-only baseline.
\noindent\textbf{Case 2 ($\lambda = 0$):} Relying solely on the alignment module increases the mean error to \textbf{54.45\%}, indicating that alignment without the source prior is unstable under severe corruption.
\noindent\textbf{Case 3 ($0 < \lambda < 1$):} Combining both components reduces the mean error to \textbf{32.08\%}, outperforming either module individually. 
These results validate that the source model provides stable predictions when the target distribution is close to the source, while the alignment module captures domain-specific shifts. Their weighted fusion effectively balances generalization and adaptation, thereby improving robustness to corruption.

\noindent \textbf{Validating the Prototype-based Alignment Module }

To evaluate the robustness of the prototype-based alignment module, we construct four disjoint subsets ($a$, $b$, $c$, and $d$), each containing 25\% of the target samples, as shown in Table~\ref{tab:ablation_disjoint_set}. Prototypes are generated from one subset while alignment is performed on another, enforcing generalization across unseen target samples. Across different prototype–alignment combinations, the proposed module consistently achieves lower error than the source-only baseline, demonstrating that the alignment mechanism effectively transfers information beyond the samples used for prototype construction.

\noindent \textbf{Effect of the Gaussian-perturbed prototype prior}
Fig.~\ref{fig:error_curve} illustrates the effect of the Gaussian-perturbed prototype prior under sparse pseudo-label settings. By constructing a class-aligned prior from unlabeled Gaussian-perturbed target predictions and using it to complete missing class prototypes, the resulting prototype memory becomes more stable for shared-subspace alignment. As a result, 3D-BBDA with the Gaussian-perturbed prototype prior achieves lower error across most corruption types, with smaller gains under background corruption.

\noindent \textbf{Calibration Analysis}
We evaluate prediction reliability using Expected Calibration Error (ECE), which measures the alignment between predicted confidence and empirical accuracy. Table~\ref{tab:result_ECE} shows that 3D-BBDA consistently yields lower ECE values than BFTT3D across corruptions, indicating improved confidence calibration under domain shifts. Better calibration enhances reliability in safety-critical applications such as robotics and autonomous systems.

\begin{table}[t]
\centering
\scriptsize
\setlength{\tabcolsep}{1.2pt}
\renewcommand{\arraystretch}{1.15}

\resizebox{\columnwidth}{!}{%
\begin{tabular}{@{}l|ccccccccccccccc|c@{}}
\toprule
\textbf{Inference}
& \rotatebox{65}{unif}
& \rotatebox{65}{gaus}
& \rotatebox{65}{backg}
& \rotatebox{65}{impul}
& \rotatebox{65}{upsamp}
& \rotatebox{65}{rbf}
& \rotatebox{65}{rbf-inv}
& \rotatebox{65}{den-dec}
& \rotatebox{65}{den-inc}
& \rotatebox{65}{shear}
& \rotatebox{65}{rot}
& \rotatebox{65}{cut}
& \rotatebox{65}{dist}
& \rotatebox{65}{oclu}
& \rotatebox{65}{lidr}
& \textbf{Mean} \\
\midrule
Standard
& 3.30 & 5.37 & 6.01 & 6.25 & 6.32
& 6.28 & 6.29 & 6.28 & 6.33 & 6.37
& 6.34 & 6.33 & 6.34 & 6.33 & 6.35
& 6.03 \\
Adapted
& 8.79 & 7.95 & 7.71 & 7.29 & 7.16
& 7.02 & 6.92 & 6.89 & 6.87 & 6.79
& 6.74 & 6.71 & 6.68 & 6.67 & 6.61
& 7.12 \\
\bottomrule
\end{tabular}%
}

\vspace{-0.4em}
\caption{Standard vs.\ adapted inference time (ms) across ModelNet40-C corruptions.}
\label{tab:inference_time}
\vspace{-1.7em}
\end{table}

\noindent \textbf{Sensitivity to noisy pseudo-labels and scalability on real-world}
Table~\ref{tab:noisy_pseudo-labels} analyzes robustness to noisy pseudo-labels. Although performance degrades as noise increases, the reduction is gradual rather than catastrophic. Even under substantial noise levels, the method maintains stable performance, suggesting resilience to imperfect pseudo-labels that may arise from real-world sensing variability.

\noindent \textbf{Computational overhead}
Table~\ref{tab:inference_time} reports the per-sample inference time,
where 3D-BBDA requires 7.12\,ms/sample compared with 6.03\,ms/sample
for standard inference, corresponding to a 1.09\,ms prediction-stage
overhead. Table~\ref{tab:runtime} further separates the additional
buffer/block-level adaptation costs from per-sample prediction.
Prototype construction is performed once for the accumulated target
buffer, while shared-subspace alignment is recomputed for each
128-sample target block. Each target sample requires one forward pass
through the frozen source model, with no backward pass or parameter
update.
\begin{table}[t]
\centering
\scalebox{0.5}{
\begin{tabular}{lcc}
\toprule
\textbf{Operation} & \textbf{Frequency} & \textbf{Time} \\
\midrule
Prototype construction
& per target buffer & 198.29 ms \\

Gaussian prior/completion
& per target condition & 350.65 ms \\

Shared-subspace alignment
& per 128-sample block & 669.26 ms \\

\midrule
Standard inference
& per sample & 6.03 ms \\

Prediction-stage inference
& per sample & 7.12 ms \\
\bottomrule
\end{tabular}
}
\vspace{-0.8em}
\caption{Runtime decomposition of 3D-BBDA on ModelNet40-C with
PointNet. Buffer/block-level adaptation costs are reported.}
\label{tab:runtime}
\vspace{-2em}
\end{table}

\noindent \textbf{Effect of Shared-Subspace Alignment}
We ablate the shared-subspace transformation $W$ by comparing
prototype-based prediction with and without $W$ as given in Table \ref{tab:w_ablation}.
On ModelNet40-C/PointNet, incorporating $W$ reduces the mean error
from 33.47\% to 32.08\%, an improvement of 1.39\%.
This confirms that shared-subspace alignment provides additional
benefit beyond prototype-based prediction alone.

\begin{figure}[t]
\centering

% ================= LEFT: FIGURE =================
\begin{minipage}[c]{0.52\columnwidth}
    \centering

    \vspace*{1.8em} % Moves the figure downward

    \includegraphics[width=\linewidth]
    {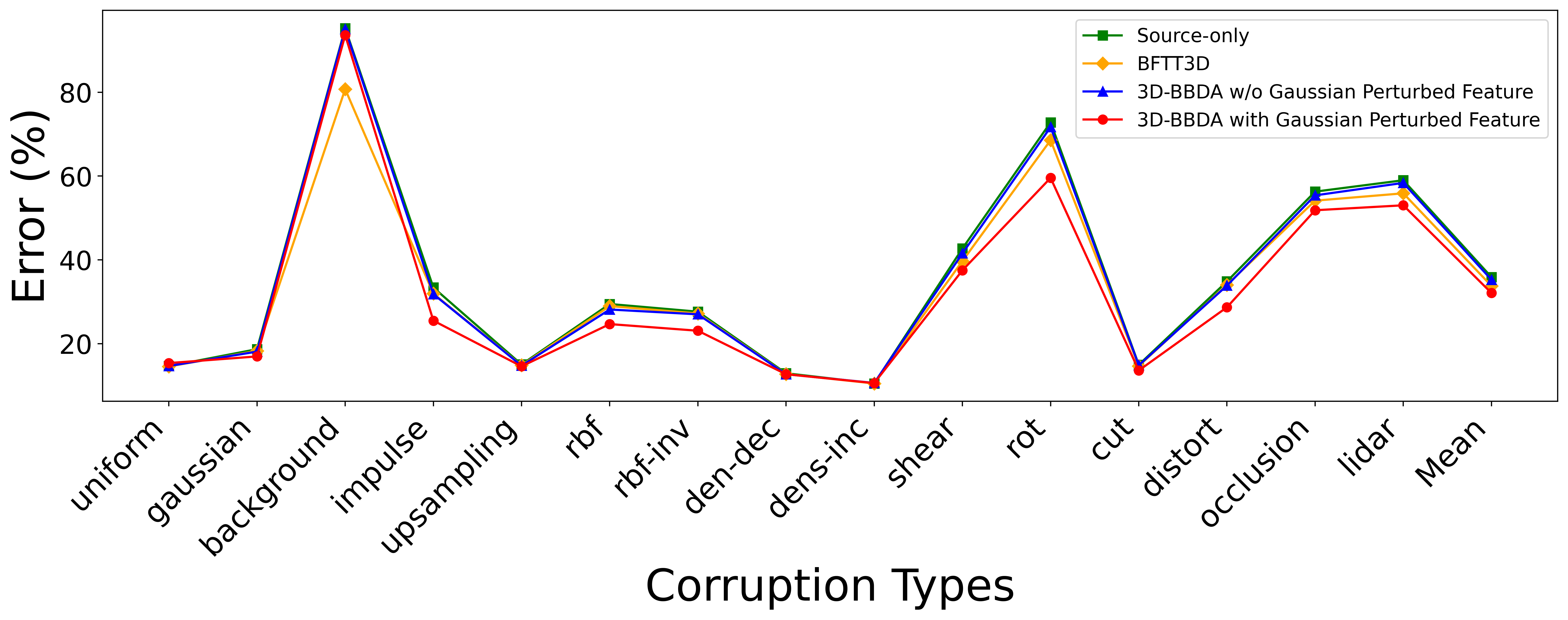}

    \vspace{-0.4em}
    \caption{Error across ModelNet-40C corruptions with
    and without the Gaussian-perturbed prior.}
    \label{fig:error_curve}
\end{minipage}%
\hfill
% ================= RIGHT: TABLE =================
\begin{minipage}[c]{0.45\columnwidth}
    \centering
    \footnotesize
    \setlength{\tabcolsep}{3pt}
    \renewcommand{\arraystretch}{1.1}

    \begin{tabular}{@{}lc@{}}
        \toprule
        \textbf{Variant} & \textbf{Error $\downarrow$} \\
        \midrule
        With $W$    & \textbf{32.08} \\
        Without $W$ & 33.47 \\
        \bottomrule
    \end{tabular}

    \captionof{table}{Effect of $W$
    (mean error, \%).}
    \label{tab:w_ablation}
\end{minipage}

\end{figure}

\section{Conclusion}
% In this paper, Black-box domain adaptation for 3D point cloud data has been addressed. Through our work, it is demonstrated that only using the samples available at the test time, a more generalized prediction could be obtained. The back-propagation-free nature of our framework mitigates the error-accumulation problem. Through the prototype-based alignment module, it can be inferred that the more generalized prototype, which simulates the source data, helps in better adaptive prediction. By leveraging the predictive capability of the source model through the weighted prediction module, for the test samples close to the source domain, better results are obtained. Exhaustive analysis proves the validity of each module integrated with our proposed approach. Our framework is capable of giving adaptive prediction in a single forward pass with low computational overhead, ensuring its relevance for real-time engineering applications such as autonomous navigation, robotic manipulation, and intelligent sensing. Future extensions will concentrate on expanding the framework to segmentation and detection tasks, in addition to integrating temporal adaptation for ongoing robotic perception. 
We proposed \textbf{3D-BBDA}, a backpropagation-free black-box domain adaptation framework for 3D point cloud classification that requires neither source data nor source model internals. The method combines pseudo-label guided prototype alignment with a Gaussian-perturbed prototype prior completion strategy to stabilize adaptation under sparse pseudo-labels, and a weighted prediction module to balance source priors with target-specific correction. Experiments on multiple corrupted 3D benchmarks demonstrate consistent robustness gains with minimal overhead, enabling real-time and privacy-preserving deployment. We validate robustness and real-time feasibility via standard corruption benchmarks and inference overhead measurements. 
% Future work will extend 3D-BBDA to 3D detection/segmentation and temporal adaptation for continuous robotic perception.
% This paper studied black-box domain adaptation for 3D point cloud classification under strict deployment constraints, where neither source data nor source model internals are accessible. We introduced \textbf{3D-BBDA}, a backpropagation-free framework that performs prototype-based alignment using only unlabeled target samples at test time. To improve stability under sparse pseudo-label spaces, we proposed a \emph{Gaussian-reference prototype completion} strategy that ensures a complete prototype memory for subsequent shared-subspace alignment. In addition, a weighted prediction module fuses the source-model logits with alignment-based logits, balancing source prior knowledge and target-specific correction while avoiding gradient updates and error accumulation. Extensive experiments across multiple corrupted 3D benchmarks demonstrate consistent robustness gains with minimal computational overhead, making 3D-BBDA suitable for real-time, privacy-preserving 3D perception in robotics and related applications. Future work will extend this framework to 3D segmentation and detection and explore temporal adaptation for continuous robotic perception.

% \input{sec/2_formatting}
% \input{sec/3_finalcopy}
\section*{Acknowledgments}
This work was supported by \textit{Indian Space Research Organisation} under Grant No.~\textit{RespondAI, P.No.ISRO/RES/3/1147/26-27}. Shivangi Rai gratefully acknowledges the financial support provided through the Visvesvaraya PhD Fellowship by the Ministry of Electronics and Information Technology (MeitY), Government of India.

{
    \small
    \bibliographystyle{ieeenat_fullname}
    \bibliography{main}
}

\clearpage

% =============================================================
% SUPPLEMENTARY MATERIAL: start on a fresh page with own title,
% and restart section/table/figure/algorithm and page numbering.
% =============================================================
% Flush any remaining two-column floats before the supplement.
\clearpage
\setcounter{page}{1}
\setcounter{section}{0}
\setcounter{table}{0}
\setcounter{figure}{0}
\setcounter{equation}{0}
\setcounter{algorithm}{0}

% Unique hyperref destinations avoid collision with main-paper labels.
\renewcommand{\theHsection}{sup.\arabic{section}}
\renewcommand{\theHtable}{sup.\arabic{table}}
\renewcommand{\theHfigure}{sup.\arabic{figure}}
\renewcommand{\theHequation}{sup.\arabic{equation}}
\providecommand{\theHalgorithm}{\arabic{algorithm}}
\renewcommand{\theHalgorithm}{sup.\arabic{algorithm}}
% Page counter intentionally restarts, as in the sample PDF. Disable
% page destinations only for the supplement to prevent duplicate anchors.
\hypersetup{pageanchor=false}
\pagestyle{plain}

% \twocolumn[<header>] produces the full-width title at the top of
% the first supplementary page, with the normal two-column text below.
\twocolumn[{%
  \vspace*{0.25in}
  \begin{center}
    {\large\bfseries
    Supplementary Material of 3D-BBDA: Backpropagation-Free Black-Box\\
    Domain Adaptation for Robust 3D Point Cloud Recognition\par}
  \end{center}
  \vspace{0.73in}
}]
\thispagestyle{plain}

\section{Detailed Comparison with BFTT3D}
\label{sec:supp_bftt}

BFTT3D is the closest prior method to 3D-BBDA because both avoid
test-time backpropagation and employ non-parametric target
representations together with prototype-based shared-subspace
alignment. The primary methodological distinction is therefore not
the introduction of a completely new subspace-alignment objective,
but the reformulation of this pipeline under a stricter source-free,
logits-only black-box interface. Table~\ref{tab:bftt_detailed}
summarizes the component-wise relationship.

The distinction in Table~\ref{tab:bftt_detailed} also clarifies which
parts of the pipeline are retained rather than newly derived.
3D-BBDA does not claim the shared-subspace alignment objective itself
as a new contribution. Instead, the method changes the information
available to this alignment: source-derived memory is removed and
replaced by pseudo-label-guided target prototypes, missing classes
are completed from an unlabeled Gaussian-perturbed target prior, and
the resulting alignment prediction is fused with the black-box source
prediction. These changes are what enable the alignment framework to
operate under the joint source-free, black-box, and
backpropagation-free constraints.
\section{Algorithm}
Algorithm~\ref{alg:3Dbbda} summarizes the complete 3D-BBDA inference
procedure. The method first constructs a fixed target-derived prototype
memory from accumulated features and pseudo-labels, then performs
block-wise shared-subspace alignment and weighted prediction fusion.

\section{When Does 3D-BBDA Provide Larger Gains?}
\label{sec:supp_gain_analysis}

The main experiments show that the magnitude of adaptation gain varies
substantially across datasets and corruption types. To better
understand this behavior, we analyze three properties of the
black-box pseudo-label stream: source confidence, predictive entropy,
and pseudo-label class coverage.  Table~\ref{tab:gain_diag_dataset} summarizes these dataset-level
diagnostics together with the corresponding adaptation gains.

For target sample $x_i^t$, let
$p_s(y\mid x_i^t)$ denote the softmax probability obtained from the
black-box source logits. We compute the mean source confidence as
\[
\mathrm{Conf}
=
\frac{1}{n_t}
\sum_{i=1}^{n_t}
\max_c p_s(c\mid x_i^t),
\]
and normalized predictive entropy as
\[
\mathrm{Ent}
=
-\frac{1}{n_t\log C}
\sum_{i=1}^{n_t}
\sum_{c=1}^{C}
p_s(c\mid x_i^t)\log p_s(c\mid x_i^t).
\]
Pseudo-label class coverage is defined as the fraction of the $C$
classes receiving at least one pseudo-labeled target sample before
Gaussian-prior completion.

\begin{table}[t]
\centering
\scalebox{0.7}{
\begin{tabular}{lcccc}
\toprule
\textbf{Dataset} &
\textbf{Confidence} &
\textbf{Entropy} &
\textbf{Coverage (\%)} &
\textbf{Gain (\%) $\uparrow$} \\
\midrule
ModelNet40-C
& 0.533 & 0.559 & 96.17 & \textbf{3.68} \\

ScanObjectNN-C
& 0.703 & 0.315 & 99.11 & 0.70 \\

ShapeNet-C
& 0.913 & 0.057 & 98.32 & 2.16 \\
\bottomrule
\end{tabular}
}
\caption{Dataset-level diagnostic statistics for the controlled
PointNet analysis. Gain is the reduction in classification error
relative to Source-only. These statistics are used only to understand
the behavior of the adaptation mechanism and do not replace the
primary benchmark results.}
\label{tab:gain_diag_dataset}
\end{table}

To obtain a finer-grained view, we pool the 43 corruption conditions
from ModelNet40-C, ScanObjectNN-C, and ShapeNet-C and measure the
Pearson correlation between the available pseudo-label diagnostics
and the adaptation gain. The resulting correlations are reported in
Table~\ref{tab:gain_corr}.
\begin{table}[t]
\centering
\begin{tabular}{lcc}
\toprule
\textbf{Diagnostic} &
\boldmath$r$ \textbf{ with gain} &
\boldmath$p$ \\
\midrule
Source confidence
& $-0.515$ & $0.0004$ \\

Predictive entropy
& $+0.471$ & $0.0014$ \\

Class coverage
& $-0.288$ & $0.0608$ \\

Missing classes
& $+0.315$ & $0.0394$ \\
\bottomrule
\end{tabular}
\caption{Correlation between black-box pseudo-label diagnostics and
the adaptation gain across 43 corruption conditions. These results are
diagnostic rather than causal.}
\label{tab:gain_corr}
\end{table}

The strongest relationship is observed for source-prediction
uncertainty. Lower source confidence is associated with larger
adaptation gains ($r=-0.515$), while higher predictive entropy is
associated with larger gains ($r=0.471$). This is consistent with
3D-BBDA being most useful when the source prediction is degraded
enough that the target-derived prototype branch can provide a
complementary correction.

Representative examples of large and near-zero adaptation gains are
reported in Table~\ref{tab:gain_examples}.

Importantly, this trend should not be interpreted as ``lower confidence
always implies better adaptation.'' The target pseudo-labels must
remain sufficiently informative to construct meaningful class
prototypes. If the black-box predictor becomes confidently wrong or
collapses its predictions onto only a subset of classes, the
target-derived prototype memory can also become biased.

\begin{table}[t]
\centering
\begin{tabular}{llc}
\toprule
\textbf{Dataset} & \textbf{Corruption} &
\textbf{Gain (\%) $\uparrow$} \\
\midrule
ModelNet40-C & Rotation     & +10.86 \\
             & Impulse      & +8.06 \\
             & Distortion   & +6.12 \\
             & LiDAR        & +5.88 \\
             & Density-inc. & +0.04 \\
\midrule
ShapeNet-C   & RBF-inv      & +4.53 \\
             & RBF          & +4.09 \\
             & Distortion   & +3.87 \\
             & Shear        & +3.78 \\
             & Density-inc. & $\approx$0 \\
\bottomrule
\end{tabular}
\caption{Examples of large and near-zero adaptation gains. Gains are
reported as the reduction in error relative to the corresponding
Source-only prediction.}
\label{tab:gain_examples}
\end{table}

SydneyUrban-C provides a useful complementary failure mode. The
black-box source predictor has high mean confidence ($0.940$) and low
predictive entropy ($0.063$), but its pseudo-label stream covers only
approximately $68.1\%$ of the 14-class label space, corresponding to
about $4.46$ missing classes on average before completion. Thus, high
confidence does not necessarily imply a high-quality pseudo-label
distribution: the predictor can be confidently concentrated on an
incomplete subset of classes.

The Gaussian prior prevents such missing classes from leaving the
prototype memory undefined, but it cannot fully recover the
class-discriminative information that is absent from the original
pseudo-label assignments. This helps explain why the improvement on
SydneyUrban-C is positive but comparatively modest.

Overall, these diagnostics suggest that 3D-BBDA benefits most from an
intermediate regime: source predictions are sufficiently degraded for
target-side correction to be useful, but remain sufficiently
informative to form meaningful pseudo-label groups. Very confident
class collapse and extremely noisy pseudo-label streams both restrict
the amount of useful corrective information available to the
prototype branch. We emphasize that this analysis establishes
empirical associations rather than a causal prediction rule for
adaptation gain.

\begin{table*}[t]
\centering
\scalebox{0.85}{
\begin{tabular}{p{3.2cm}p{5.5cm}p{6.5cm}}
\toprule
\textbf{Aspect} &
\textbf{BFTT3D} &
\textbf{3D-BBDA (Ours)} \\
\midrule

\textbf{Problem setting}
&
Backpropagation-free 3D test-time adaptation.
&
Source-free, logits-only black-box, and backpropagation-free
3D adaptation.
\\

\textbf{Core intuition}
&
Construct a prototype memory and align target representations with
this memory without gradient-based parameter updates.
&
Retain prototype-based alignment while replacing source-dependent
adaptation memory with a memory constructed entirely from unlabeled
target data and black-box predictions.
\\

\textbf{Source data during adaptation}
&
Required for constructing the source-derived adaptation memory.
&
Not required.
\\

\textbf{Source-model interface}
&
Uses source-derived feature/prototype information.
&
Frozen source model is accessed only through its final
predictions/logits.
\\

\textbf{Test-time backpropagation}
&
No.
&
No.
\\

\textbf{Target representation}
&
Non-parametric target representation.
&
Frozen PointNN representation independent of the source-model
architecture.
\\

\textbf{Prototype source}
&
Source-derived feature memory.
&
Unlabeled target features grouped according to black-box
pseudo-labels.
\\

\textbf{Prototype construction}
&
Prototype/memory information originates from the source domain.
&
For each pseudo-class, accumulated target features assigned to that
class are averaged to construct the class prototype.
\\

\textbf{Missing-class prototypes}
&
No explicit target-only completion mechanism.
&
Missing pseudo-label classes are completed using the
Gaussian-perturbed target prototype prior.
\\

\textbf{Shared-subspace alignment}
&
Aligns source-derived prototype memory with target features.
&
Retains the shared-subspace alignment principle, but replaces
source-derived memory with completed target-derived prototypes.
$W$ is estimated from target-derived prototypes and target features.
\\

\textbf{Role of source samples in alignment}
&
Source information enters through the source-derived adaptation
memory.
&
No source samples or source features participate in the alignment;
the alignment depends only on $n_p$ target-derived prototypes and
target features.
\\

\textbf{Prediction from alignment}
&
Prototype/alignment similarity provides the adapted prediction.
&
Prototype similarities provide alignment logits $l_{am}$ after
target-only prototype completion and shared-subspace projection.
\\

\textbf{Use of source prediction}
&
No explicit weighted source/alignment logit fusion.
&
Black-box source logits are retained and fused with alignment logits:
$l_{\mathrm{final}}
=\lambda l_s+(1-\lambda)l_{am}$.
\\

\textbf{Robustness to incomplete pseudo-label coverage}
&
Relies on the source-derived prototype memory.
&
Gaussian completion explicitly maintains a class-complete prototype
memory when target pseudo-label assignments are incomplete.
\\

\textbf{Adaptation memory}
&
Source-derived memory whose size depends on the stored source
representations.
&
One completed target-derived prototype per class; memory therefore
grows linearly with $C$.
\\

\textbf{Memory entries ($C=55$)}
&
8,907.
&
55 (\textbf{161.9$\times$ fewer}).
\\

\textbf{Alignment-system size ($C=55$)}
&
9,035.
&
183.
\\

\textbf{Architecture/interface dependence}
&
Requires the source-derived information needed to construct its
prototype memory.
&
Source architecture is opaque to the adaptation mechanism; only its
final predictions are consumed.
\\

\textbf{Contribution boundary}
&
Backpropagation-free prototype-based 3D test-time adaptation.
&
Retains the prototype/shared-subspace alignment principle but
reformulates it for strict source-free black-box deployment through
target-only prototype construction, Gaussian missing-class completion,
and source/alignment prediction fusion.
\\

\bottomrule
\end{tabular}
}
\caption{\textbf{Detailed methodological comparison between BFTT3D
and 3D-BBDA.} 3D-BBDA retains the prototype-based shared-subspace
alignment principle but removes source-derived adaptation memory.
The memory and alignment-system sizes correspond to ShapeNet-C with
$C=55$ and PointNet.}
\label{tab:bftt_detailed}
\end{table*}

\section{Details of Controlled Recent Baseline Ports}
\label{sec:supp_baselines}

Recent adaptation methods differ substantially in the information
they assume available at test time. We therefore distinguish native
implementations from controlled ports and explicitly document the
access assumptions used for each comparison.
Table~\ref{tab:baseline_constraints} summarizes the information-access
and architecture assumptions of these recent baselines.
\begin{table*}[t]
\centering
% \scriptsize
% \setlength{\tabcolsep}{5pt}
\scalebox{0.8}{
\begin{tabular}{lccccc}
\toprule
\textbf{Method} &
\textbf{Source data} &
\textbf{Source internals} &
\textbf{Target BP} &
\textbf{Architecture-specific} &
\textbf{Evaluation type} \\
\midrule

3D-BBDA
& No
& No (logits only)
& No
& No
& Native \\

PURGE-GATE PG-SF
& No
& CLS/Q/K
& No
& Transformer
& Native method \\

PURGE-GATE PG-SP$^\dagger$
& Source-derived statistics
& Point features
& No
& Controlled port
& PointNet/GDANet \\

DINE-3D$^\ddagger$
& No
& Source probabilities only
& Yes
& No
& Controlled 2D$\rightarrow$3D port \\

LN-TTA
& No
& LN/token representations
& No
& Transformer
& PointTransformer only \\
\bottomrule
\end{tabular}
}
\caption{Information-access assumptions of the recent baselines.
$^\dagger$ denotes our controlled PointNet/GDANet port of
PURGE-GATE PG-SP. $^\ddagger$ denotes our controlled feature-space
baselinesport of the 2D black-box adaptation method DINE.}
\label{tab:baseline_constraints}
\end{table*}

\subsection{Controlled PURGE-GATE \cite{yazdanpanah2025purge} Port}

Native PURGE-GATE PG-SF is designed for Transformer point-cloud
architectures and relies on CLS and attention Q/K representations.
These quantities are not available in PointNet or GDANet and, more
importantly, would not be accessible through the logits-only
black-box interface assumed by 3D-BBDA.

For the PointNet/GDANet comparison, we therefore construct a
controlled PG-SP$^\dagger$ port. Purging is applied to the earliest
64-dimensional point-wise representation available in each backbone,
and source-derived embedding statistics are used according to the
PG-SP formulation. No target-side backpropagation is performed.

\begin{table}[t]
\centering
\scalebox{0.8}{
\begin{tabular}{llccc}
\toprule
\textbf{Dataset} &
\textbf{Backbone} &
\textbf{PG-SP$^\dagger$} &
\textbf{DINE-3D$^\ddagger$} &
\textbf{3D-BBDA} \\
\midrule

ModelNet40-C
& PointNet & 36.18 & 46.53 & \textbf{32.08} \\
& GDANet   & 32.61 & 46.72 & \textbf{29.62} \\

ScanObjectNN-C
& PointNet & 60.46 & 68.29 & \textbf{59.48} \\
& GDANet   & 65.71 & 72.27 & \textbf{65.64} \\

ShapeNet-C
& PointNet & 36.57 & 60.90 & \textbf{32.74} \\
& GDANet   & 24.38 & 54.74 & \textbf{22.85} \\

SydneyUrban-C
& PointNet & 73.19 & 77.82 & \textbf{72.42} \\
\bottomrule
\end{tabular}
}
\caption{Mean corruption error (\%, $\downarrow$) for the controlled
PURGE-GATE and DINE ports.}
\label{tab:supp_recent_baselines}
\end{table}

\subsection{Controlled 2D$\rightarrow$3D DINE \cite{liang2022dine} Port}

DINE was originally proposed for domain adaptation from black-box
2D predictors. To evaluate whether a direct transfer of a 2D BBDA
objective is sufficient for the 3D corruption setting, we construct
DINE-3D$^\ddagger$ as a controlled feature-space port.

For comparability, DINE-3D uses the same frozen PointNN target
representation used by 3D-BBDA. We retain the principal optimization
components of DINE, including source-probability distillation,
EMA-based prediction memory, information-maximization objectives,
MixUp, and target-side fine-tuning. Consequently, unlike 3D-BBDA,
DINE-3D requires backpropagation through the target-side adaptation
model.
As shown in Table~\ref{tab:supp_recent_baselines}, the direct DINE
port does not transfer favorably to these 3D corruption benchmarks.
For example, on ModelNet40-C/PointNet its mean error is 46.53\%,
compared with 35.88\% for Source-only and 32.08\% for 3D-BBDA.
A similar degradation is observed across the remaining
dataset/backbone combinations.

We emphasize that this experiment should not be interpreted as
showing that the original DINE method is intrinsically unsuitable for
3D data. DINE-3D$^\ddagger$ is a controlled feature-space port
designed to answer a narrower question raised by the reviewers:
whether directly transferring an established 2D black-box adaptation
objective to the same 3D target representation is sufficient. Under
this controlled transfer, the answer is negative.

\subsection{Comparison with LN-TTA \cite{yazdanpanah2026revisiting}}
\label{sec:supp_lntta}

LN-TTA is a recent backpropagation-free test-time adaptation method
designed for Transformer-based point-cloud architectures. Its
adaptation mechanism relies on LayerNorm-related representations and
Transformer-specific tokens, including the CLS token. In contrast,
the PointNet and GDANet backbones used in our main experiments are
BatchNorm-based architectures and do not provide these
Transformer-specific representations.

Consequently, LN-TTA cannot be applied unchanged under our
PointNet/GDANet evaluation protocol. A numerical comparison would
require either replacing the source backbone with a Transformer or
modifying LN-TTA to operate on representations for which its original
adaptation mechanism was not defined. Such a comparison would therefore
conflate the adaptation method with a change in backbone or constitute
a non-equivalent reimplementation.

We therefore do not include LN-TTA in the main quantitative tables.
Instead, we explicitly report this architectural incompatibility.
This distinction also highlights the difference in deployment
assumptions: LN-TTA requires access to specific internal
LayerNorm/token representations, whereas 3D-BBDA treats the frozen
source model as a black box and accesses only its final prediction
outputs.

\begin{algorithm}[!t]
    \small
    \caption{3D-BBDA inference procedure.}
    \label{alg:3Dbbda}
    
    \textbf{Input}: Source-trained black-box predictor $F_S$,
    unlabeled target-domain data $\{x_i^t\}_{i=1}^{n_t}$,
    non-parametric network $F_{nn}$.\\
    \textbf{Output}: Predicted target labels
    $\{y_i^t\}_{i=1}^{n_t}$.

    \begin{algorithmic}[1]

    \STATE Extract target features $\{f_{t,i}\}_{i=1}^{n_t}$ using
    $F_{nn}$ (Eqs.~(1)--(5)).

    \STATE Query $F_S$ once for each target sample to obtain source
    logits $l_{s,i}$ and pseudo-labels $\hat{y}_i^t$.

    \STATE Accumulate target features and pseudo-labels and construct
    class prototypes $\{P_i\}$ using Eq.~(6).

    \STATE Complete missing class prototypes using the
    Gaussian-perturbed prototype prior (Eqs.~(7)--(8)).

    \STATE Partition the target samples into alignment blocks of
    128 samples.

    \FOR{each target alignment block}
        \STATE Compute the shared-subspace transformation $W$ using
        the fixed prototype memory and current target block
        (Eqs.~(11)--(13)).
        
        \STATE Compute prototype similarities and alignment logits
        $l_{am}$ (Eqs.~(9)--(10)).
        
        \STATE Fuse source and alignment logits using Eq.~(14):
        \[
        l_{\mathrm{final}}
        = \lambda l_s + (1-\lambda)l_{am}.
        \]
    \ENDFOR

    \STATE \textbf{Return} predicted target labels
    $\{y_i^t\}_{i=1}^{n_t}$ from $l_{\mathrm{final}}$.

    \end{algorithmic}
\end{algorithm}

% {
%     \small
%     \bibliographystyle{ieeenat_fullname}
%     \bibliography{main}
% }

% % Supplementary section numbering: S1, S2, S3, ...
% \setcounter{section}{0}
% \renewcommand{\thesection}{S\arabic{section}}

% \section*{Supplementary Material}

% \input{supplementary}

\end{document}